\documentclass[sigconf]{acmart} %, review
\usepackage[ruled,vlined]{algorithm2e}
\usepackage{bbm}
\usepackage{multirow}
\usepackage{makecell}

\newcommand{\ud}[1]{\underline{#1}}
\newcommand{\udb}[1]{\underline{\bf{#1}}}

\AtBeginDocument{%
  }

\copyrightyear{2023}
\acmYear{2023}
\setcopyright{acmlicensed}
\acmConference[KDD '23] {Proceedings of the 29th ACM SIGKDD Conference on Knowledge Discovery and Data Mining}{August 6--10, 2023}{Long Beach, CA, USA.}
\acmBooktitle{Proceedings of the 29th ACM SIGKDD Conference on Knowledge Discovery and Data Mining (KDD '23), August 6--10, 2023, Long Beach, CA, USA}
\acmPrice{15.00}
\acmDOI{10.1145/3580305.3599817} 
\acmISBN{979-8-4007-0103-0/23/08}
\begin{document}

%%
%% The "title" command has an optional parameter,
%% allowing the author to define a "short title" to be used in page headers.
\title{Evolve Path Tracer: Early Detection of Malicious Addresses in Cryptocurrency}

%%
%% The "author" command and its associated commands are used to define
%% the authors and their affiliations.
%% Of note is the shared affiliation of the first two authors, and the
%% "authornote" and "authornotemark" commands
%% used to denote shared contribution to the research.

%%
%% By default, the full list of authors will be used in the page
%% headers. Often, this list is too long, and will overlap
%% other information printed in the page headers. This command allows
%% the author to define a more concise list
%% of authors' names for this purpose.
% \renewcommand{\shortauthors}{Trovato and Tobin, et al.}

\author{Ling Cheng}
\email{lingcheng.2020@phdcs.smu.edu.sg}
\affiliation{%
  \institution{Singapore Management University}
  \country{Singapore}}
\author{Feida Zhu}
\authornote{Corresponding author}
\email{fdzhu@smu.edu.sg}
\affiliation{%
  \institution{Singapore Management University}
  \country{Singapore}}
\author{Yong Wang}
\email{yongwang@smu.edu.sg}
\affiliation{%
  \institution{Singapore Management University}
  \country{Singapore}}
\author{Ruicheng Liang}
\email{rcliang@mail.hfut.edu.cn}
\affiliation{%
  \institution{Hefei University of Technology}
  \city{Heifei}
  \country{China}}
\author{Huiwen Liu}
\email{hwliu.2018@phdcs.smu.edu.sg}
\affiliation{%
  \institution{Singapore Management University}
  \country{Singapore}}

\renewcommand{\shortauthors}{Cheng et al.}

%%
%% The abstract is a short summary of the work to be presented in the
%% article.
\begin{abstract}
   With the boom of cryptocurrency and its concomitant financial risk concerns, detecting fraudulent behaviors and associated malicious addresses has been drawing significant research effort. 
   Most existing studies, however, rely on the full history features or full-fledged address transaction networks, both of which are unavailable in the problem of early malicious address detection and therefore failing them for the task. 
   To detect fraudulent behaviors of malicious addresses in the early stage, we present \emph{Evolve Path Tracer} 
   which consists of \emph{Evolve Path Encoder LSTM}, \emph{Evolve Path Graph GCN}, and \emph{Hierarchical Survival Predictor}.
   Specifically, in addition to the general address features, we propose Asset Transfer Paths and corresponding path graphs to characterize early transaction patterns. 
   Furthermore, since transaction patterns change rapidly in the early stage, 
   we propose \emph{Evolve Path Encoder LSTM} and \emph{Evolve Path Graph GCN} to encode asset transfer path and path graph under an evolving structure setting.
   \emph{Hierarchical Survival Predictor} then predicts addresses' labels with high scalability and efficiency.
   We investigate the effectiveness and generalizability of \emph{Evolve Path Tracer} on three real-world malicious address datasets. 
   Our experimental results demonstrate that \emph{Evolve Path Tracer} outperforms the state-of-the-art methods. 
   Extensive scalability experiments demonstrate the model's adaptivity under a dynamic prediction setting.
\end{abstract}

%%
%% The code below is generated by the tool at http://dl.acm.org/ccs.cfm.
%% Please copy and paste the code instead of the example below.
%%


%% Keywords. The author(s) should pick words that accurately describe
%% the work being presented. Separate the keywords with commas.

%% A "teaser" image appears between the author and affiliation
%% information and the body of the document, and typically spans the
%% page.
%\begin{teaserfigure}
%  \includegraphics[width=\textwidth]{sampleteaser}
%  \caption{Seattle Mariners at Spring Training, 2010.}
%  \Description{Enjoying the baseball game from the third-base
%  seats. Ichiro Suzuki preparing to bat.}
%  \label{fig:teaser}
%\end{teaserfigure}

%%
%% This command processes the author and affiliation and title
%% information and builds the first part of the formatted document.
\maketitle

%------------------------------------------------------------------------
\section{Introduction}
\label{sec:intro}
The past decade has witnessed the growth of cryptocurrency as decentralized global financial systems. 
Unfortunately, it has long been criticized for accommodating various cyber-crimes due to its anonymity. 
According to the recent Crypto Crime Report by Chainalysis\footnote{https://go.chainalysis.com/2023-Crypto-Crime-Report.html}, 
2022 was the biggest year ever for crypto-crime, with $3.8$ billion dollar worth of crypto asset stolen. Researchers and practitioners have made significant efforts to combat fraudulent activities and identify  associated \emph{malicious addresses}, particularly for Bitcoin (BTC) due to its singularly important leadership remained unshakable among all cryptocurrencies.

%Among all cryptocurrencies, Bitcoin (BTC) has much scarcer on-chain address data than other platforms (e.g., ETH, EOS with smart contracts).
%Also, most BTC-based methods are compatible with other cryptocurrency platforms.
%Thus, researchers and practitioners have made significant efforts to fight against these fraudulent activities and identify the associated \emph{malicious addresses} on BTC.
%
%
\begin{figure}
	\centering
	\vspace{-0ex}
	\includegraphics[width=1.0\columnwidth, angle=0]{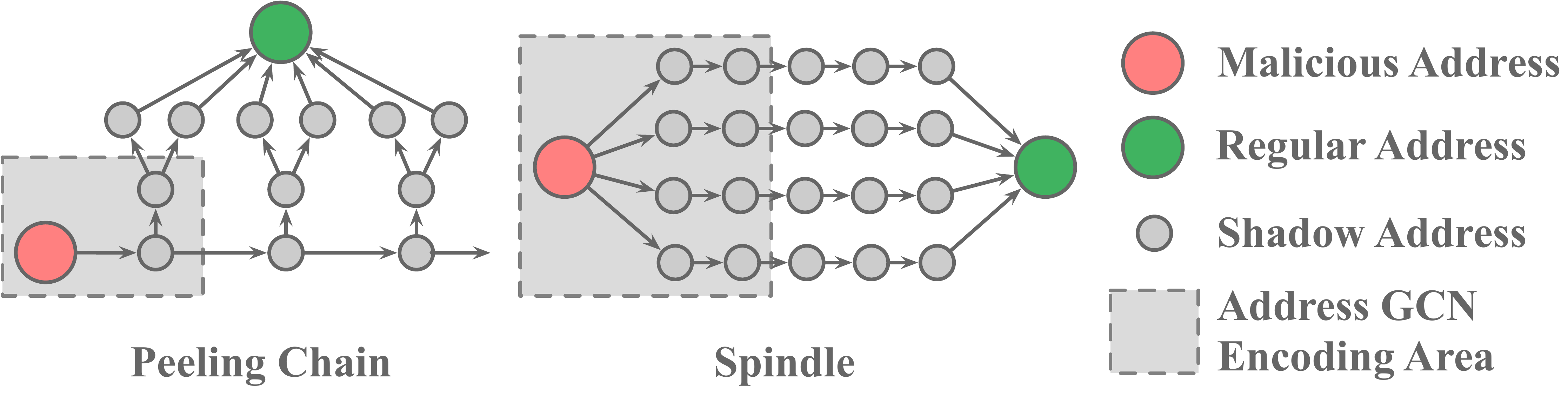}
	\vspace{-4ex}
	\caption{Two asset flow patterns of malicious addresses. 
                 Address transaction network methods may suffer shadow address issues as they only consider neighbors within 2 or 3 hops.
                 By encoding asset transfer paths (from red node to green node), the model can capture more comprehensive patterns.
                 }
	\label{fig:intro_motivation}
	\vspace{-4ex}
\end{figure}
%
%
% About Existing Works
% Most existing detection methods~\cite{10_, 2_, 11_, 12_, 15_} focus on designing features to characterize specific types of malicious activity with detailed case studies.
% Besides, by combining statistic analysis and visualization technology, they successfully identified some representative malicious transaction patterns.
% Recent studies \cite{5_,6_,7_} further leverage deep-learning techniques to detect malicious addresses. 
% By encoding account features and the transaction network structure with deep-learning models, they achieve great performance improvement in malicious address detection.
% What previous works haven't done
However, there are three major limitations in current methods:
\begin{itemize}
    \item \textbf{Ineffective for early detection}. Most previous works adopt random walk and graph neural network~\cite{13_, 14_, 7_, 6_} to improve the performance of general malice detection. But most of them require a full-fledged address network which is unavailable under the early stage settings, as early transaction graphs are usually unconnected and fast-evolving, and most addresses are unlabelled.

    \item \textbf{Type-specific features are not generalizable}. Malice are evolving fast. It is impossible to build a unique feature set for every new malicious activity. Most methods~\cite{8_, 9_, 10_} are designed for specific malicious behaviors and cannot be generalized to other types and other cryptocurrencies. Although some studies \cite{31_} can detect categories of entities, the features are too general to describe malicious transaction patterns.

    \item \textbf{Challenge of shadow account issues}. Most address transaction network methods may suffer the shadow address issues. As shown in Fig.~\ref{fig:intro_motivation}, the perpetrators create a bunch of shadow addresses,\footnote{https://www.elliptic.co/blog/elliptic-analysis-bitcoin-bitfinex-theft}\footnote{Using Blockchain Analysis to Mitigate Risk} and the illicit asset will be transferred through the paths constructed by these addresses and eventually converge at a particular destination. Usually, these paths are extremely long compared to the encoding area (2 or 3 hops) of address network methods~\cite{7_}, as such address networks cannot encode transaction patterns comprehensively. If expanding the encoding area, they may incur scalability issues and run into the problem of minority class dilution.
\end{itemize}

% Why our work is significant
Considering the first limitation, in this work, we first set up the \emph{Early Malicious Address Detection (EMAD)} task, which is urgent but seldom discussed by existing studies. 
Then we design a novel model \emph{Evolve Path Tracer} to detect malicious addresses at an early stage.
For the second requirement, considering malicious addresses' objective is to transfer illicit money to a legal place. Even though the behavior patterns differ for various address types, we can still derive their intentions by monitoring their asset transfer patterns. 
We therefore develop a path extractor to extract those significant \emph{Asset Transfer Paths} to characterize the early-stage transactions of various malicious addresses effectively. 

For the last requirement, illicit activities are usually organized together for specific purposes, and the behavior patterns evolve fast during the early stage. As a result, encoding each path individually may miss critical information.
Therefore, after encoding paths independently with path encoder LSTM, we build an asset path GCN module to encode the inter-relation among paths. 
In this graph, asset transfer paths connect to one other if they share the same intersection addresses.
As shown in Fig.~\ref{fig:intro_motivation}, by encoding paths with the same destination together, 
we can discover the hidden intention of malicious addresses and address the tricky problem of shadow addresses challenging most address GCN models.
Moreover, static encoding models cannot capture the dynamism of fast-evolving early transfer patterns. This problem is similar to the one indicated by \cite{28_}, we therefore embed the evolving mechanism into our path encoder LSTM and path GCN module for more sophisticated path representations under the dynamic setting.

Considering the scalability issue in the real-world application, we implement a \emph{Hierarchical Survival Prediction} module to alleviate the workload of feature preparation during the prediction. 
Previous prediction results can be directly used in the next time step, which empowers the model with a faster prediction speed and the ability to deal with a dynamic setting.
%
% Contribution
% We identify for our work the following contributions. 
In summary, the contributions of this paper can be summarized as follows:
\begin{itemize}
    \item
    The Asset Transfer Paths are proposed for the \emph{EMAD} task, which is urgent but seldom discussed. 
    These paths exhibit high versatility in monitoring transaction patterns of various malicious behaviors in the early stage. 
    This novel concept can be potentially applied to all current cryptocurrency systems. 
    
    \item
    We propose the \emph{Evolve Path Tracer} model that can fully utilize the asset transfer paths to dynamically encode various transaction patterns with path encoder LSTM. 
    Besides, to relieve the problem of shadow addresses, the model can also encode the paths' structural relationship under a dynamic setting with a novel evolve path graph GCN module. 
    The versatility and flexibility are unachievable by other existing models. 
    
    \item
    We conduct extensive evaluations to assess the model's effectiveness, and the results show that \emph{Evolve Path Tracer} deliver a substantially better performance for three different illicit address datasets than the state-of-the-art models. 
    Also, owing to the \emph{Hierarchical Survival Prediction} module, our \emph{Evolve Path Tracer} can effectively predict addresses' labels and scales well for increasing data.
\end{itemize}

We organize the remaining sections as follows. 
In Section~\ref{sec:related_work}, we classify and review related works.
In Section~\ref{sec:prob_define}, we define the problem of Early Malicious Address Detection.
Then we explain the construction of the asset transfer path and path graph in Section Section~\ref{sec:method_path}, 
and introduce the details of Evolve Path Tracer in Section~\ref{sec:method_encoder}. 
Finally, we present experimental results in Section~\ref{sec:experiment}
and conclude this paper in Section~\ref{sec:conclusion}.

%------------------------------------------------------------------------
\section{Related Work}
\label{sec:related_work}
%Early detection of malicious addresses on cryptocurrency is an urgent yet seldom discussed task.
The methods for malice detection on cryptocurrencies can be classified into three categories, namely case analysis, machine learning, and graph based methods.
We review these methods according to their types. 

\subsection{Case Analysis} Case analysis mainly focuses on addresses' behaviors in a certain case.
Reid et al. \cite{8_} identified entities by considering similar transaction times over an extended timeframe.
Androulaki et al. \cite{9_} considered several features of transaction behavior, including the transaction time, the index of addresses, and the value of transactions.
Jourdan et al. \cite{10_} explored five types of features, including address features, entity features, temporal features, centrality features, and motif features.
Vasek et al. \cite{2_} gave a list of Bitcoin scams and conducted a statistical study.
Case-Related features are often helpful in interpreting certain cases based on heuristic clustering and tainted fund flow.
However, these methods require intensive case analysis, and most of the insights are only available in some specific cases, let alone apply to other platforms with complex heterogeneous relationships in general~\cite{kdd_rv_3, kdd_rv_8}.

\subsection{Machine Learning} 
Machine Learning methods can automatically learn general address features to increase model's generalization ability.
Yin et al. \cite{11_} applied supervised learning to classify entities that might involve in cybercriminal activities.
Akcora et al. \cite{12_} applied the topological data analysis (TDA) approach to generate the bitcoin address graph for ransomware payment address detection. 
Shao et al. \cite{15_} embedded the transaction history into a lower-dimensional feature for entity recognition.
Nerurkar et al. \cite{31_} used several general features for classifying different address categories.
The general address features improve the model's generality significantly. 
However, they are challenging to characterize addresses' behaviors comprehensively. 
Particular transaction patterns of asset flow are difficult to be reflected in these characteristics.
Besides, some models~\cite{36_,37_,38_} can detect crypto-malware early, 
TLR model~\cite{review_1} learns the expected likelihood score from training data and uses the disparity between predicted and observed likelihood scores.
while they require malware's operation logs or code corpus, which are unavailable in most cases where we only have on-chain data.

\subsection{Graph Based Methods}
Graph-based methods focus on the interaction patterns between object addresses and related addresses.
Harlev et al. \cite{13_} considered transaction graph features to predict entity types.
Wu et al.\cite{14_} proposed two types of network motifs to detect BTC mixing service addresses.
Weber et al.~\cite{7_} encodes address transaction graph with GCN, Skip-GCN, and Evolve-GCN to detect illicit addresses.
Chen et al.\cite{6_} proposed E-GCN for phishing node detection on the ETH platform.
Tam et al.\cite{16_} proposed EdgeProp to learn the large-scale transaction network representations for illicit account identification.
Lin et al.\cite{17_} proposed random walk-based embedding methods to encode specific network features for node classification.
By changing the sampling strategy, Wu et al.\cite{29_} proposed the Trans2Vec model for a similar task.
Li et al.\cite{30_} encoded the temporal information of historical transactions for phishing detection.
Chen et al.\cite{kdd_rv_9} proposed the AntiBenford subgraph framework for anomaly detection in the Ethereum.
Network-based methods perform well for retrospect analysis, as they encode the structural information. 
Besides, most graph-based anomaly detection can also be potentially implemented. 
AMNet~\cite{review_2} and BWGNN~\cite{review_3} aim to discriminate anomaly nodes with the spectral energy distribution difference.
However, in the early stages, the performance degrades greatly if the networks are of sparse structures for the emerging networks with few connections\cite{kdd_rv_10}. 
Also, due to the limitation of scalability, these methods suffer shadow address issues as mentioned in Fig.~\ref{fig:intro_motivation}. 
Expanding the encoding area may lead to Over-Smoothing issues and the dilution of the minority class~\cite{34_} under the data-unbalanced setting~\cite{kdd_rv_5, kdd_rv_6, kdd_rv_7}.

Among all cryptocurrencies, Bitcoin (BTC) has the largest volume, and is considered as the prototype for other cryptocurrency platforms (e.g., ETH, EOS with smart contracts),
most BTC-based methods are thus compatible with other cryptocurrency platforms.
Following previous practitioners, we also focus on detecting \emph{malicious addresses} on BTC.

% \subsection{Early Rumor Detection}
% Cho et al.~\cite{27_} used GRU, a typical neural network for sequence modeling. At each time split, previous hidden state and current summation features are fed into the GRU unit to predict the labels for the given samples.
% Song et al.~\cite{25_} proposed CED, which also uses GRU for sequence modeling. They proposes the concept of “Credible Detection Point,” to increase the prediction speed.
% Yuan et al.~\cite{23_} developed a multi-source long-short term memory network (M-LSTM) to model user behaviors by using a variety of user edit aspects as inputs.
% Zheng et al.~\cite{22_} put forward SAFE. Instead of predicting the labels directly, it generates hazard rates for the survival models. 
% Lv et al.~\cite{35_} built a transformer-based multi-modal feature representations to capture the multi-level dependencies among multi-modal content.

%------------------------------------------------------------------------
\section{Problem Definition}
\label{sec:prob_define}
% Addresses and transactions are the two major attributes of each Bitcoin transaction record. 
% Wallet addresses result from a series of digital signature operations based on the public key, calculated from the private key with the elliptic encryption algorithm.

For each BTC transaction ($TX$), we analyze its input TX set $I$= $\{i_1, i_2, \dots, i_{|I|}\}$ and an output TX set $J$= $\{ j_1, j_2, \dots, j_{|J|}\}$. 
% If the funds in this transaction are unspent, then the output will be marked as UTXO (unspent transaction output). 
% A 
Each $tx$ records token distribution between $I$ and $J$. 
Narratively speaking, the incoming tokens flow into a pool and then flow to the outgoing transactions according to the prior agreement proportion
\footnote{https://www.walletexplorer.com/txid/520a1edf88f9afc4a6dba554a952f98911388aabf1f\\7648ad5e71b2ae8b5d5e4}. 
% Thus, there is no record of how many tokens an input transaction $i$ gives to an output transaction $j$. 
There is no record of how many tokens flow from an Input $i$ to an Output $j$.
We thus have to build a complete transaction bipartite graph for this $TX$
% . 
% As a result, we 
and generate $|I|\times|J|$ transaction pairs in total. 
In other words, a transaction has $|I|\times|J|$ transaction pairs inside.
If an address acts as the input address of $TX$,
we denote $TX$ as the address's $Spend$ transaction.
Otherwise, $Receive$ transaction.

By the $t_m$-th time step, 
$D_{t_m}$=$\{d^i_{t_m}\}_{i=1}^N$=$\{(l^i, T_{in, t_m}^{i}, T_{out, t_m}^{i})\}_{i=1}^N$ is the input data,
where $l^i \in \{0, 1\}$ is the label of Address $i$, and $0$ or $1$ stands for regular and malicious addresses respectively.
$T_{in, t_m}^{i}$=$[TX_{in, 1}^{i}, \dots, TX_{in, N_{in,t_m}}^{i}]$ are the $Spend$ transaction sets 
and $T_{out, t_m}^{i}$=$[TX_{out, 1}^{i},\dots, TX_{out, N_{out,t_m}}^{i}]$ 
are the $Receive$ transaction sets of Address $i$ by the $t_m$-th time step respectively.

\vspace{0.2cm}
\noindent\textbf{Early Malicious Address Detection (EMAD).}
Given a set of addresses $A$, and 
$D_{t_m}$ at timestep $t_m$, % = $\{(Add^i, T_{in, t_m}^i, T_{out,t_m}^i)\}_{i=1}^N$, 
the problem is to build a binary classifier $F$ such that 
\begin{equation}
F(d^i_{t_m})=
\begin{cases}
1& \text{if Address $i$ is malicious}\\
0& \text{Otherwise}
\end{cases}.
\end{equation}
In the early detection task, to prevent the model predicts conflict labels at different timesteps, which will confuse users, we thus require the prediction to be consistent and predict the correct label as early as possible.
We denote the confident time as $t_{c}$, where all classifier predictions $F$ after $t_{c}$ are consistent. 
The smallest $t_{c}$ is denoted as $t_{f.c}$. Our purpose is to train a classifier that predicts the correct label of an address with a smaller $t_{f.c}$.

%------------------------------------------------------------------------
\section{Asset Transfer Path and Path Graph}
\label{sec:method_path}
As we know the destination and the source of asset transfer can provide critical information,
we thus propose the \emph{Asset Transfer Paths} and \emph{Asset Transfer Path Graph} to reflect: 
1) the characteristics of each path, 
2) the interaction between paths, 
and 3) the characteristics of the asset source (destination). 

Take Address $i$'s $Recieve$ transaction set as an example. 
For a $Receive$ transaction in the set, 
suppose we find its significant asset source, which is also a transaction, 
then we link them to form an asset transfer pair. 
After we trace the asset source iteratively, the asset transfer pairs form an asset transfer path naturally.
If all paths come from the same source, then we can say that, at a particular time, 
there are multiple transactions initiated at the same time, and all these transactions' destinations are Address $i$.
If we encode each path as a node, these nodes can be connected through this source of funds, thus forming a graph.

\begin{figure}
	%\centering
	\vspace{-0ex}
	\includegraphics[width=1.\columnwidth, angle=0]{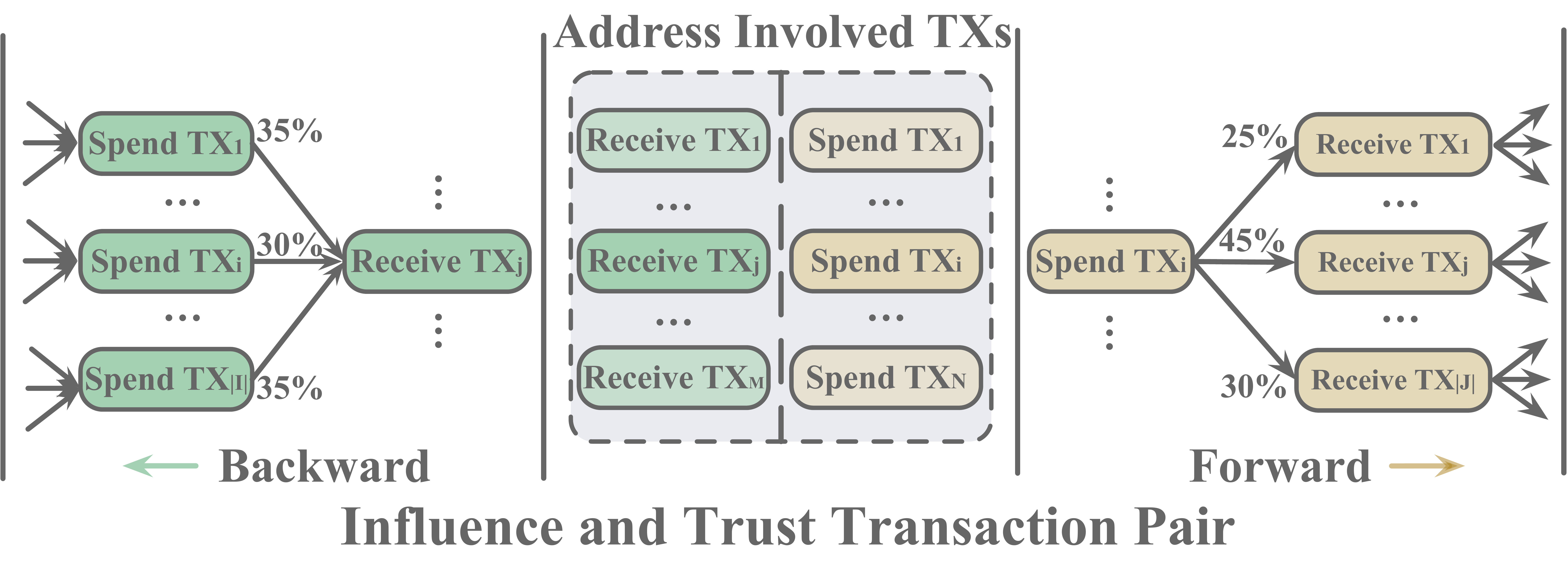}
	\vspace{-6ex}
	\caption{
Asset transfer pairs. Given an address, we collect its transaction history (All its Receive 
 and Spend TXs in the dotted box).
 For each Receive TX, we trace its asset source from the inflow (Green Spend TXs) to build Influence TX pairs.
 For each Spend TX, we trace its asset destination from the outflow (Yellow Receive TXs) to build Trust TX pairs. $N=N_{in,t_m}$, $M=N_{out,t_m}$.}
	\label{fig:transaction_pair}
	\vspace{-4ex}
\end{figure}

\subsection{Transaction Pair and Path Construction}
Not all transaction pairs help identify illicit addresses. Those noteworthy pairs typically constitute a significant amount portion of the entire transaction.
%We call such transactions \emph{significant transactions}, and we would only include such significant transactions in this study to prune the most unimportant ones. 
%
As illustrated in Fig.~\ref{fig:transaction_pair}, $TX_j$ receives assets from three spend TXs (each contributing $35\%$, $30\%$, and $35\%$ to the total transaction amount). 
On the other hand, the spend $TX_i$ transfers out BTCs to three receive TXs (with a distribution of $25\%$, $45\%$, and $30\%$ as in this example). Given an Address $i$ and a time step $t_m$, 
TXs in the shaded dotted-lined box represent all spend TXs, and receive TXs occurred up to timestep $t_m$ associated with Address $i$.

As mentioned in Section \ref{sec:prob_define}, given a set $I$= $\{i_1, i_2, \dots i_{|I|}\}$ of $|I|$ spend transactions to an receive transaction $j$ and the set $\{{I\rightarrow j}\}$ of all 
% (Input TX, Output TX) 
transaction pairs, i.e., $\{{I\rightarrow j}\}$=$\{(i_1,j),(i_2,j),\ldots,(i_{|I|},j)\}$, we define \emph{Influence Transaction Pair} as follows: 
Given an influence activation threshold $\theta$,
$(i_{k},j)$ is called an \emph{Influence Transaction Pair} for transaction $j$,
if the amount of transaction pair $(i_{k},j)$ ($1\le k\le |I|$) contributes at least a certain proportion of the received amount of transaction $j$, i.e, $\hat{A}(i_{k},j) \ge \theta \times \hat{A}(\{{I\rightarrow j}\})$ where $\hat{A}(\cdot)$ denotes the amount of a transaction pair or the sum of all transaction pairs.

Similarly, given a set $J$= $\{j_1, j_2, \dots j_{|J|}\}$ of $|J|$ transactions, and a transaction $i$, and the set $\{{i\rightarrow J}\}$ of all transaction pairs whose spend transaction is $i$, i.e., $\{{i\rightarrow J}\}$=$\{(i,j_1),(i,j_2),\ldots,(i,j_{|J|})\}$.
If there exists a receive transaction $j_k$ ($1\le k\le |J|$), such that transaction $i$ transfers at least a certain proportion of its spend amount to it, this transaction pair is called a \emph{Trust Transaction Pair} for transaction $i$, as it indicates a certain form of trust from $i$ to $j_k$ in terms of asset transfer. 

To construct asset transfer paths, given an influence transaction pair $(i_{k},j)$, we can conclude that transaction $j$ obtains at least a significant amount (based on the threshold) of the asset in this transaction from transaction $i_{k}$.
% It follows that given a transaction $j$. 
%
Accordingly, given a transaction $j$, if there exists a sequence of transaction pairs such that 
(I) each pair is an influence transaction pair; 
(II) the spend transaction of each pair is the receive transaction of the previous pair; 
and (III) the receive TX of the last pair is transaction $j$, we call such a sequence an \emph{Backward Path} for $j$ as indicated by the green arrow in Fig. \ref{fig:transaction_pair}. It reveals where $j$ obtains the asset and can be used to trace back to the source of the asset.
The detail to prepare the backward asset transfer path is shown in Algorithm~\ref{alg:bk_path}. 
Similarly, we can define a \emph{Forward Path} to trace the destinations of transaction $i$'s asset flow. 
For brevity, we would refer to both the \emph{Backward Path} and \emph{Forward Path} as \emph{Asset Transfer Paths}. 

As shown in Fig.~\ref{fig:asset_transfer_path_and_graph}, for Receive Tx R-1 in the Flow of Receive Tx, 
after we trace its asset source, we can get three critical transfer pairs, namely (10 $\rightarrow $R-1, 11 $\rightarrow $R-1, 12 $\rightarrow $R-1). 
As shown in \emph{Backward Asset Transfer Path}, iteratively, we can get four paths (P-1, P-2, P-3, P-5) ended with R-1.

\subsection{Asset Transfer Path Graph}
\label{sec::Asset_Transfer_Path_Graph}
\begin{figure}
	%\centering
	\vspace{-0ex}
	\includegraphics[width=1.\columnwidth, angle=0]{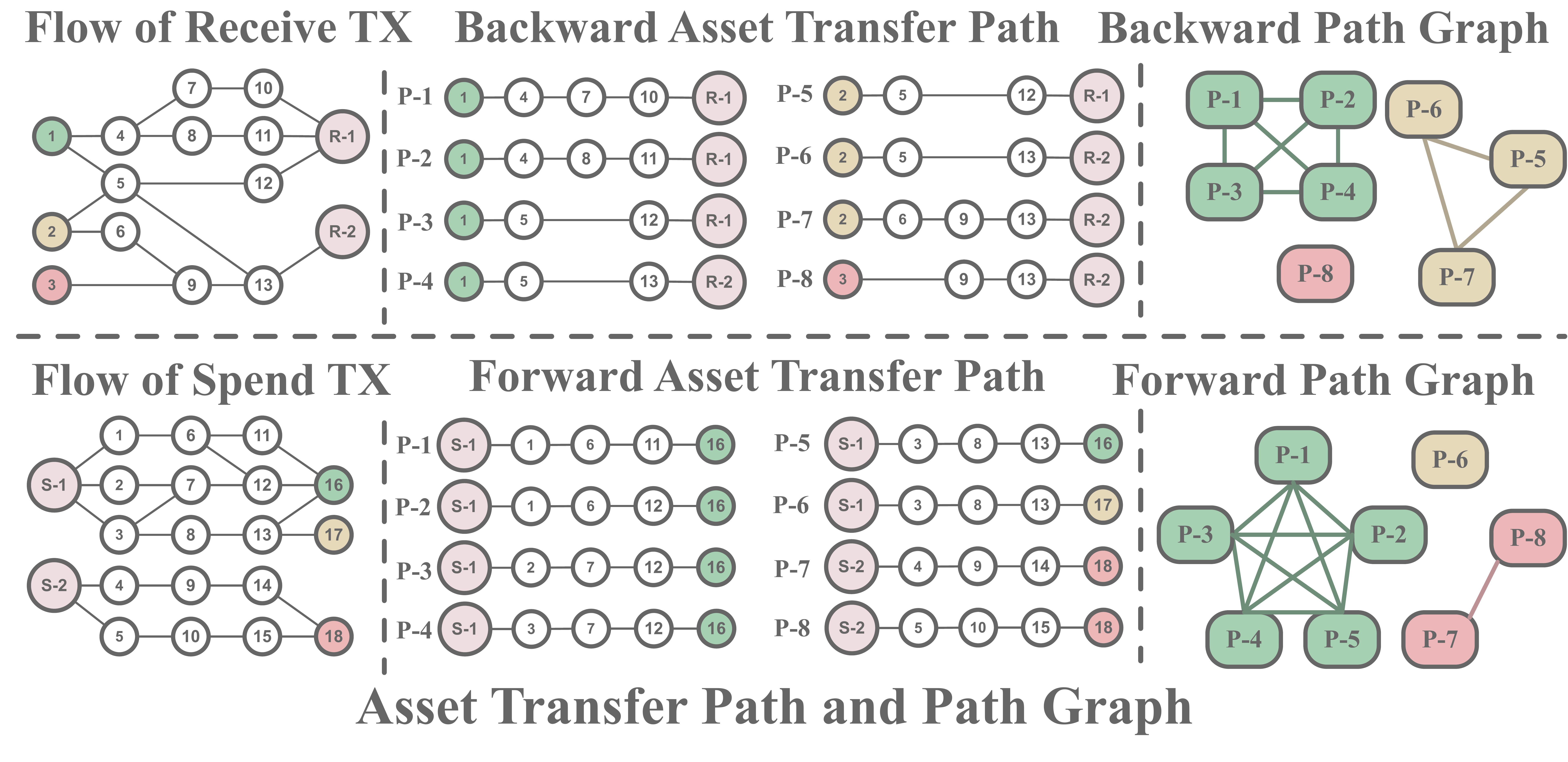}
	\vspace{-4ex}
	\caption{
Asset transfer path and path graph. 
 By tracing the asset source iteratively, we get a series of Influence TX pairs. We combine them end-to-end to form Backward Asset Transfer Path (Similar to Forward Asset Transfer Path). 
 If paths have the same source or destination, we connect them through their intersection to form a path graph (Graph in the same color). 
 Different colors stand for different starting or ending points.}
	\label{fig:asset_transfer_path_and_graph}
	\vspace{-4ex}
\end{figure}

The address transaction graph methods may suffer from the perturbation of shadow addresses and the scalability issues caused by mixing services. 
However, even though malicious addresses use mixing services, their suspicious asset transfer paths will still converge. 
Therefore, unlike the previous address-based graph, we build graphs based on the asset transfer path. 
As shown in Fig. \ref{fig:asset_transfer_path_and_graph}, in the path graph part, each node represents an asset-transfer path. 
If two paths share the same source (for backward paths) or destination (for forward paths), we then connect them with an edge, and we thus can get a group of fully-connected graphs. 
Here we also take \emph{Backward Asset Transfer Paths} as examples. Among them, three paths(P-1, P-2, P-3) have the same source (Tx-1 colored green).
Also, another path (P-4) ended with R-2 is also initiated by the same source,
we thus group them into a path graph, which is colored green, as shown in the \emph{Backward Path Graph}.

Since every source or destination has a binding address at a specific timestep, 
we use the feature of this binding address at this time point to represent the edge feature in the corresponding graph.
The Address Features (AF) and transaction features are elaborated in Appendix.

%------------------------------------------------------------------------
\section{Evolve Path Tracer}
\label{sec:method_encoder}
\begin{figure}
	%\centering
	\vspace{-0ex}
	\includegraphics[width=1.\columnwidth, angle=0]{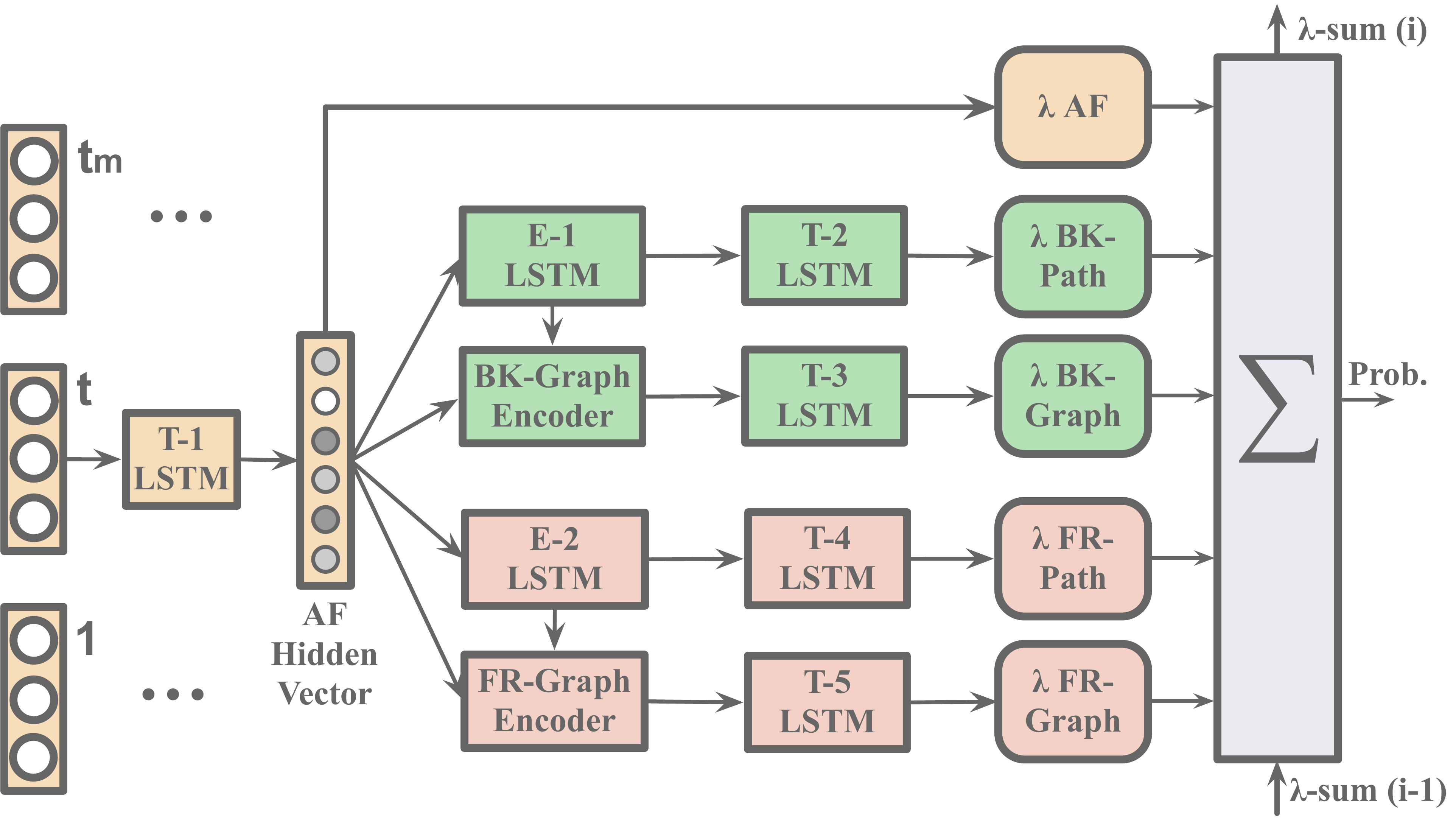}
	\vspace{-3ex}
	\caption{Detailed pipeline of \emph{Evolve Path Tracer}. 
	At each timestep, five Temporal-LSTM models (T LSTM) encode temporal information of different features. 
        The Evolve Path Encoder LSTM (E LSTM) and Evolve Path Graph GCN (Graph Encoder) will be updated to encode asset transfer paths and path graphs dynamically. For module detail description, please refer to Table~\ref{tab:module_explain}.}
	\label{fig:model_detail}
	\vspace{-4ex}
\end{figure}

As shown in Fig.~\ref{fig:model_detail}, at the $t$ th timestep,
five Temporal-LSTM models (T-1 to T-5 LSTM) are implemented to encode temporal information of the following input data:
\textbf{1}: Address Feature (AF), 
 \textbf{2}: Backward Path (BK-Path), 
 \textbf{3}: Backward Graph (BK-Graph), 
 \textbf{4}: Forward Path (FR-Path), 
 \textbf{5}: Forward Graph (FR-Graph).

Take the backward branch as an example (green branch), we input the current AF to update T-1 LSTM.
However, as we have multiple backward paths at each timestep, to update other LSTMs, we need to encode each path and aggregate them as the inputs for these LSTMs.
Therefore, we implement E-1 LSTM (Backward Evolve Path Encoder LSTM) to encode each backward path into a path vector.
Meanwhile, to reflect the path structure's dynamism, the parameters of E-1 LSTM are calculated based on the current ($t$ th) hidden vector of T-1 LSTM and the previous ($t-1$ th) hidden vector of T-2 LSTM.
With an attention-weighted summation, all these backward path vectors are aggregated as the input to update T-2 LSTM.
Besides, BK-Graph Encoder (Backward Evolve Path Graph GCN) incorporates path-graph structure in these backward path vectors.
Then, following similar processings, these structure-aware path vectors are aggregated as the input to update T-3 LSTM.
The forward branch follows the same mechanism to update T-4 and T-5 LSTMs.
Concretely, at the $t$ th timestep, the process can be summarized in five steps:
\begin{itemize}
    \item \textbf{Step-1}. To encode the temporal information of the Address Feature(AF), we input current ($t$ th timestep) AF into T-1 LSTM to get the T-1 hidden vector.
    \item \textbf{Step-2}. We concatenate the T-1 hidden vector with the previous ($t-1$ th timestep) T-2 and T-3 hidden vectors to generate parameters for E-1 LSTM and BK-Graph Encoder.
    \item \textbf{Step-3}. E-1 LSTM encodes each path into a corresponding path vector. Then we weighted sum these path vectors to update T-2 LSTM.
    \item \textbf{Step-4}. BK-Graph Encoder updates path vectors with graph information. Similarly, we weighted sum these vectors to update T-3 LSTM.
    \item \textbf{Step-5}. After similar processing for the forward branch, we use the hidden vectors of five temporal LSTM (T-1 to T-5) to predict the current label of the address.
\end{itemize}

\subsection{Evolve Path Encoder LSTM}
\label{sec:path_encoder}
% AF-LSTM / T-1 LSTM
The Address Feature is the basis for modeling the address transaction pattern. 
We implement an LSTM (\emph{T-1 LSTM}) to encode Address Feature's temporal information and guide the processing in other modules.
\begin{equation}
\label{eq:T_1_LSTM}
h_{t}^{T_1}, c_{t}^{T_1} = \mbox{LSTM}^{T_1}(f^u_t, h_{t-1}^{T_1}, c_{t-1}^{T_1}),\\
\end{equation}
where $h_{t}^{T_1} \in \mathbb{R}^{d}$ and $c_{t}^{T_1} \in \mathbb{R}^{d}$ are the hidden state and the cell state of \emph{T-1 LSTM} at time $t$. 
$f^u_t \in \mathbb{R}^{d_n}$ is the address feature at time $t$. 
$d$ and $d_n$ are the dimensions of the hidden state and address feature.

% Path Encoding Evolve LSTM / E-1 LSTM
As mentioned in Section \ref{sec:method_path}, the asset transfer path is composed of a series of transaction nodes. The lengths of these paths are different. To encode them uniformly, we project an original path $P_{ori}$ to a uniform path $P_u$ with the length of $L_u$. Given an original path $P_{ori}$ with length of $L_{ori}$, the zoom ratio is calculated by $R_z$ = $L_{ori}/L_u$, then the $i$-th (start from $0$) node in $P_u$ is calculated by the average feature of the ($\lfloor i\times{R_z} \rfloor$)-th node to the ($\lceil (i+1)\times{R_z} \rceil$-1)-th node of $P_{ori}$, where $\lfloor \cdot \rfloor$ and $\lceil \cdot \rceil$ are round down and round up functions. Then, we denote the uniform path as:
\begin{equation}
\label{eq:Asset_Transfer Path}
\mathit{P_{u}} = [p_1, p_2, \dots, p_{L_u}],
\end{equation}

Path structures are different on different addresses and timesteps.
We may lose dynamic information with a static encoder. 
In Evolve-GCN~\cite{28_}, the weights are updated by previous weights or those representative nodes but may dismiss the individual address property. 
Instead, to update the parameters of a specific Path Encoder LSTM module, we generate the new weights with the concatenation of $h_{t}^{T_1}$
and previous hidden state of corresponding Temporal LSTM. 
For the $j$-th node in the input asset transfer path, backward \emph{Evolve Path Encoder LSTM} (\emph{E-1 LSTM}) computes the following function:
\begin{equation}
\begin{aligned}
\label{eq:E_1_LSTM_Path_Enocder_LSTM}
% H_t^{p} &= [h_{t}^{T_1} || h_{t-1}^{T_2}], \\
%i_j &= \sigma((W_{ii}H_t^{p}){p_j} + (W_{hi}H_t^{p}){h_{j-1}^{E_1}} + b_{i}H_t^{p}), \\
%f_j &= \sigma((W_{if}H_t^{p}){p_j} + (W_{hf}H_t^{p}){h_{j-1}^{E_1}} + b_{f}H_t^{p}), \\
%g_j &= \mbox{tanh}((W_{ig}H_t^{p}){p_j} + (W_{hg}H_t^{p}){h_{j-1}^{E_1}} + b_{g}H_t^{p}), \\
%o_j &= \sigma((W_{io}H_t^{p}){p_j} + (W_{ho}H_t^{p}){h_{j-1}^{E_1}} + b_{o}H_t^{p}), \\
%c_j &= f_j\odot{c_{j-1}} + i_j\odot{g_j}, \\
%h^{E_1}_j &= o_j\odot{\mbox{tanh}(c_j)},
h^{E_1}_j, c^{E_1}_j = \mbox{LSTM}^{E_1}([h_{t}^{T_1} || h_{t-1}^{T_2}], {h_{j-1}^{E_1}}, {c_{j-1}^{E_1}}),
\end{aligned}
\end{equation}
where $||$ stands for concatenation, $h_{t-1}^{T_2} \in \mathbb{R}^{d}$ is the hidden state of \emph{T-2 LSTM} at timestep $t$-$1$. 
\emph{T-2 LSTM} encodes the temporal information of the backward asset transfer path set.
Besides, all parameters in the \emph{E-1 LSTM} are represented by the multiplication of $[h_{t}^{T_1} || h_{t-1}^{T_2}]$ and corresponding learnable weights.
% $W_{**} \in \mathbb{R}^{d\times{d}\times{2d}}$ and $b_{*} \in \mathbb{R}^{d\times{2d}}$ are learnable weights that transfer $H_t^{p}$ to the weights of projection layers and bias terms.
% $\sigma$ stands for sigmoid function, $\odot$ is the Hadamard product.
% 
Finally, each backward asset transfer path is denoted as the final hidden state $h^{E_1}_{L_u}$ of \emph{E-1 LSTM}. %$f^{p}$ which equals to the 
The representation of the $i$-th path at timestep $t$ is $f^{p}_{i, t}$. 
% In this way, asset transfer paths can capture addresses' individual characteristics and their temporal features.

% Path-Attention and Update Path Global LSTM T-2
For path aggregation, we expect to select more informative paths, 
we thus adopt multi-head attention for the selection.
\begin{equation}
a_{i, t}^{j} = W^{a, j}\mbox{tanh}(W^{p, u}[f^p_{i, t} || h_{t}^{T_1}]); \quad {\alpha}_{i, t}^{j} = \mbox{Softmax}(a_{i, t}^{j}), 
\end{equation}
%\begin{equation}
%{\alpha}_{i, t}^{j} = \mbox{Softmax}(a_{i, t}^{j}) = \mbox{exp}(a_{i, t}^{j})/{\sum_{k=1}^{N_{E_1}}\mbox{exp}(a_{k, t}^{j})},
%\end{equation} 
\begin{equation}
\hat{f}^{p}_{t} = ||_{j=1}^{{M}^{p}} \hat{f}^{p, j}; \quad \hat{f}^{p, j}_{t} = \sum_{i=1}^{N_{E_1}}{\alpha}_{i, t}^{j}f^{p}_{i, t},
\end{equation}
where $f^{p}_{i, t}$ is the representation of the $i$-th path at timestep $t$. 
$W^{p,u} \in \mathbb{R}^{\frac{d}{M^p}}\times{2d}$ and $W^{a, j} \in \mathbb{R}^{{1}\times{\frac{d}{M^p}}}$ are learnable matrices, 
$j$ stand for the index of the attention head, $M^p$ is the total head number. $N_{E_1}$ is the backward asset transfer path number.
where $\hat{f}^{p, j}_{t}$ is the weighted summed path feature vector of $j$-th head,
$\hat{f}^{p}_{t}$ is the concatenation of all heads' output. 
The hidden state $h_{t}^{T_2}$ and cell state $c_{t}^{T_2}$ of \emph{T-2 LSTM} are updated as:
\begin{equation}
h_{t}^{T_2}, c_{t}^{T_2} = \mbox{LSTM}^{T_2}(\hat{f}^{p}_{t}, h_{t-1}^{T_2}, c_{t-1}^{T_2}).
\end{equation}

%------------------------------------------------------------------------
\subsection{Evolve Path Graph GCN}
\label{sec:graph_encoder}
% The interactions between each pair of paths are also beneficial for modeling and reasoning the address transaction patterns. 
If several paths are initiated by or converge at the same transaction, it may indicate certain suspicious patterns.
By encoding the relationships between these paths, the model can capture critical transaction patterns.
Also, due to the volatility of the path graph, we may lose the discriminative characteristics with a static model. 
To resolve this challenge, we propose \emph{Evolve Path Graph GCN}.
% where the model's weights are proposed by the combination of temporal address feature and the pre-step graph feature.
Take backward asset transfer paths as an example, the nodes in the path graph are updated as follows:
\begin{equation}
\begin{aligned}
H_t^{g} &= [h_{t}^{T_1} || h_{t-1}^{T_3}], \\
f^{g}_{t} &= \sigma(\mathcal{\tilde{D}}^{-\frac{1}{2}}\mathcal{\tilde{A}}\mathcal{\tilde{D}}^{-\frac{1}{2}}(f^{p}_{t}W^{g}H_t^{g})), \\
\tilde{\mathcal{A}} &= \mathcal{A} + \mathcal{I}; \quad \mathcal{A}_{i,:,j} = (W^{e}H_t^{g})S_{i,j}, \\
\mathcal{\tilde{D}} &= \mbox{diag}(\sum_{j}(A_{i,j}+\mathcal{I}_{i,j})),
\end{aligned}
\end{equation}
where $h_{t-1}^{T_3} \in \mathbb{R}^{d}$ is the hidden state of \emph{T-3 LSTM} at time $t$-$1$.
\emph{T-3 LSTM} encodes the temporal information of the backward path graph. 
$f^{p}_{t} \in \mathbb{R}^{{N_{E_1}\times{d}}}$ are the representations of path set at timestep $t$.
$A \in \mathbb{R}^{{N_{E_1}}\times{d}\times{N_{E_1}}}$ and $\mathcal{I} \in \mathbb{R}^{{N_{E_1}}\times{d}\times{N_{E_1}}}$
are the adjacent matrix and the identity matrix respectively.
If the $i$-th path and $j$-th path have the same source, 
then $A_{i,:,j}$=$\mathbf{1} \in \mathbb{R}^d$. Otherwise, $A_{i,:,j}$=$\mathbf{0} \in \mathbb{R}^d$.
If the $i$-th path and $j$-th path have the same source, 
$S_{i,j} \in \mathbb{R}^{d_n}$ is the intersection address feature of path $i$ and path $j$.
Otherwise, $S_{i,j}$=$\mathbf{0} \in \mathbb{R}^{d_n}$.
$W^{g} \in \mathbb{R}^{{d}\times{d}\times{2d}}$ and $W^{e} \in \mathbb{R}^{{d}\times{d_n}\times{2d}}$ are learnable weights, 
and they project $H_t^{g}$ to the weights of the corresponding projection layers. 
So $\mathcal{A}\in \mathbb{R}^{{N_{E_1}}\times{d}\times{N_{E_1}}}$.

% Path-Attention and Update Path Global LSTM T-3
We denote the output of \emph{Evolve Path Graph GCN} encode as interaction-aware path vectors.
Resemble the previous calculation, we adopt multi-head attention to select significant signals from these interaction-aware path vectors.
We denote the $\hat{f}^{g}_{t}$ as the final result of multi-head attention of interaction-aware path vectors. 
The hidden state $h_{t}^{T_3}$ and cell state $c_{t}^{T_3}$ of \emph{T-3 LSTM} are updated as:
\begin{equation}
h_{t}^{T_3}, c_{t}^{T_3} = \mbox{LSTM}^{T_3}(\hat{f}^{g}_{t}, h_{t-1}^{T_3}, c_{t-1}^{T_3}).
\end{equation}

%------------------------------------------------------------------------
\subsection{Hierarchical Survival Predictor}
\label{sec:survival_module}
Due to the property of consistent prediction, survival analysis \cite{22_} is proved to be effective in the early detection task. 
The survival function $S(t)$ of an event represents the probability that this event has not occurred by time $t$.
The hazard rate function $\lambda_{t}$ is the event's instantaneous occurrence rate at time $t$ given that the event does not occur before time $t$. 
In our case, the observation time is discrete in our case, we use $t$ to denote a timestamp.
The association between $S(t)$ and $\lambda_{t}$ can be calculated as:
\begin{equation}
\begin{aligned}
\label{eq:dsicrete_survival_function}
S(t)&=P(T\geq{t}) = \sum_{k=t}^{\infty}f(x), \\
\lambda_{t}&=f(t)/S(t); \quad S(t) = \mbox{exp}({-\sum_{k=1}^{t}\lambda_{k}}).
\end{aligned}
\end{equation}
Considering the model's scalability during the real-time prediction, 
we define the event as ``the address is benevolent'' and we call hazard rate as benevolent rate.
As the majority addresses are negative (benevolent),
if the address is classified as benevolent, we remove it from the monitoring list to reduce the computation cost.

To get more consistent predictions, previous work~\cite{22_} deployed a \emph{Softplus} function ${\lambda}_{t}({x_t}) = ln(1+\mbox{exp}({x_t}))$ to guarantee the hazard rate ${\lambda}_{t}$ is always positive.
Hence, the survival probability $S(t)$ monotonically decreases. 
However, the model can hardly classify addresses correctly in the early hours.
Some false predictions will never be corrected with the monotonically decreasing survival probability.
Therefore, we release this restriction with a $tanh$ activation function for benevolent rate calculation.
The consistency is assured by \emph{Consistency Loss Function} which will be discussed later.

We designed five parallel benevolent rates corresponding to each kind of information (address feature, path feature (backward and forward), and graph feature (backward and forward)). 
At time step $t$, the calculation of these benevolent rates and the prediction are as follows:
\begin{equation}
\begin{aligned}
\lambda_{j, t} &= \mbox{tanh}({W^{hz}_{T_j}h_{t}^{T_j}}),\\
\hat{y}^t &= \mbox{exp}(-\mbox{ReLU}(\sum_{i=1}^{t}\sum_{j=1}^{5}\lambda_{j, i})),
\end{aligned}
\end{equation}
where $W^{hz}_{T_j} \in \mathbb{R}^{1\times{d}}$ is the linear projection matrices for the output of \emph{T-j LSTM}.
At each time step, survival analysis first sums all previous benevolent rates, then it sums the current $5$ benevolent rates hierarchically.
Once addresses' current benevolent rates reach a certain threshold, 
we can remove them from the monitoring list to speed up the prediction and relieve the computing cost in the following hours.

%------------------------------------------------------------------------
\subsection{Training and Dynamical Prediction}
\label{sec:train_predict}
\noindent\textbf{Model Training}.
The model should give higher $S(t)$ to malicious addresses and lower $S(t)$ to benevolent addresses in every time split.
For Address $i$, at timestep $t_m$, the early detection likelihood function and the negative logarithm prediction $loss^P$ are shown as below:
\begin{equation}
\begin{aligned}
likelihood & = (1-S(t_m))^{1-l_i}S(t_m)^{l_i}\\
& = (1-\mbox{exp}(-\sum_{t=1}^{t_m}\sum_{j=1}^{5}\lambda_{j, t}))^{1-l_i}(\mbox{exp}(-\sum_{t=1}^{t_m}\sum_{j=1}^{5}\lambda_{j, t}))^{l_i},\\
loss^P_{i,t_m} & = {l_i}\sum_{t=1}^{t_m}\sum_{j=1}^{5}\lambda_{j, t} + {(l_i-1)}ln(1-\mbox{exp}(-\sum_{t=1}^{t_m}\sum_{j=1}^{5}\lambda_{j, t})).
\end{aligned}
\end{equation}
Besides, $loss^P$ is weighted by $\sqrt{t_m}$ to avoid the perturbation in the early period due to the data insufficiency.

\noindent\textbf{Consistency-boosted Loss Function}.
Since the rate function is not guaranteed to be positive in our model, a consistency loss $loss^C$ is necessary for consistent predictions. 
In every time split, the benevolent rate should have the same sign as the previous time split.
\begin{equation}
loss^C_{i,t_m} =
\begin{cases}
0& \mbox{sign}(\lambda_{t_m-1}*\lambda_{t_m})>=0 \\
1& \mbox{else}
\end{cases},
\end{equation}
where $\lambda_{t_0} = 0$. Similarly, the model should be able to rectify the poor prediction in the early period, the $loss^C$ is thus also weighted by $\sqrt{t_m}$.
%
% Besides, since the numbers of positive and negative instances are imbalanced, different penalty coefficients are allocated to each class. 
% Then, given a set of training samples with $N_p$ malicious addresses and $N_n$ legal addresses, the overall loss function is defined as:
% \begin{equation}
% \begin{aligned}
% \mathscr{L} = \sum_{t=1}^{t_M}\sqrt{t}(&C^{+}\sum_{i=1}^{N_p}(loss^P_{i,t} + \gamma{loss^C_{i,t}})+\\
%                                &C^{-}\sum_{i=1}^{N_n}(loss^P_{i,t} + \gamma{loss^C_{i,t}})),
% \end{aligned}
% \end{equation}
% where $C^{+}$ and $C^{-}$ are inversely proportional to the number of positive and negative instances in our settings. 
% $\gamma$ is a coefficient to control the contribution between $loss^P$ and $loss^C$. 
%
The overall loss function is then defined as:
\begin{equation}
\mathscr{L} = \sum_{t=1}^{t_M}\sum_{i=1}^{N}\sqrt{t}(loss^P_{i,t} + \gamma{loss^C_{i,t}}),
\end{equation}
where N is the dataset size, $\gamma$ is a coefficient to control the contribution between $loss^P$ and $loss^C$.

\begin{figure}
	%\centering
	\vspace{-0ex}
	\includegraphics[width=1.\columnwidth, angle=0]{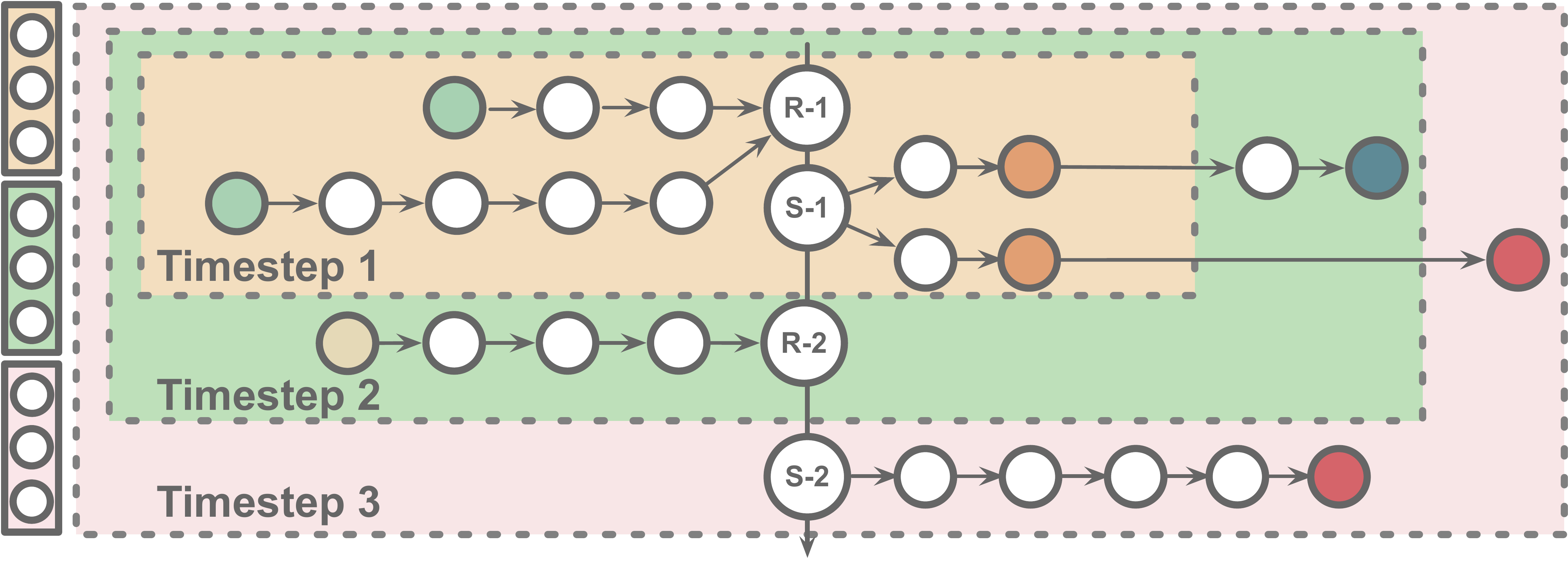}
	\vspace{-3ex}
	\caption{Dynamical Construction of asset transfer path. The three vectors on the left are Address Features corresponding to Timestamp $1$ to Timestamp $3$. Different dash boxes represent input information at different timestamps.}
	\label{fig:dynamic_construction}
	\vspace{-2ex}
\end{figure}

\noindent\textbf{Dynamical Prediction}.
Besides the ``Early Stop'' mechanism provided by \emph{Hierarchical Survival Predictor}, 
our dynamical construction scheme of asset transfer paths can also relieve the time cost of feature preparation.
As shown in Fig.~\ref{fig:dynamic_construction}, the path data can be reused if no new transaction occurs in this interval.
If an address has new \emph{Receive} or \emph{Spend} transactions, the model will create new backward or forward asset transfer paths accordingly.
Moreover, the model also checks the endpoint of the forward paths to determine whether they need to be extended or not.

%------------------------------------------------------------------------
%\subsection{Dynamical Prediction}
%\label{sec:loss_func}
%\input{section/5_4_Dynamical_Prediction}

%------------------------------------------------------------------------
\section{Experiment and Analysis}
\label{sec:experiment}
In this section, we perform empirical evaluation to answer the following research questions:
\begin{itemize}
\item \noindent\textbf{RQ1}: What is the performance with respect to different uniform path lengths?
\item \noindent\textbf{RQ2}: Does Evolve Path Tracer outperform the state-of-the-art methods for early malicious address detection? 
\item \noindent\textbf{RQ3}: How does each component benefit the final detection performance?
\item \noindent\textbf{RQ4}: Does the time overhead of data preparation and the model's scalability satisfy the real-time requirement?
\end{itemize}

\subsection{Data Collection and Preparation}
\label{sec:data}
% \noindent\textbf{Raw Data and Label Collection}
The transaction data are publicly accessible by running a Bitcoin client. 
We obtained all the data from the $1$-st to the $700,000$-th block for higher credibility, as we only collect addresses verified by enough participants. For a given address, we get the related transaction history based on the APIs exposed by BlockSci~\cite{18_}.
Based on this transaction history, we can calculate the related features and prepare the asset transfer paths and path graphs.
More detail about the label acquisition and the statistical properties of the asset transfer path can be found in Appendix~\ref{sec:appendix}.
%
%
% \noindent\textbf{Negative Address Collection}
% Since we can only get a small number of labeled illicit (positive) addresses with no verified labeled negative addresses. 
% To deal with this problem, the Positive and Unlabeled (PU) learning model was commonly used in this setting. 
% %
% For each illicit address, we also sampled some unlabeled addresses from those blocks where this illicit address conducted transactions.
% According to \cite{21_}, 15\% positive addresses, as well as the unlabeled addresses, are labeled as negative addresses.
% % set with negative labels. 
% These 15\% positive addresses are also known as spy instances.
% %
% Generally, spy instances have a higher probability of being predicted as positive than unlabeled addresses.
% Thus, we set the probability threshold $\theta$ as the value that can maximize the difference in the growth rates between the cumulative proportion of unlabeled instances and spy instances. 
% Addresses with a probability lower than $\theta$ will be regarded as reliable negative instances.
%
Table.~\ref{tab:addr_num} shows the summarized descriptions: 
The definition and numbers of positive (Posi), negative (Nega), and Positive/Negative ratio (P/N) for each malicious type (H: Hack, R: Ransomware, D: Darknet).
\begin{table}
    \caption{Dataset Statistics}
    \vspace{-0.3cm}
    \begin{center}
    \fontsize{10}{12}\selectfont
    \renewcommand{\arraystretch}{1.2}
    \begin{tabular}{ccccc}
    \toprule
    \toprule
    Type  & Definition & Posi. & Nega. & P/N(\%) \cr
    \midrule
    H & Hack and steal tokens           & 302     & 6582   & 4.03\cr
    R & Encrypt data for ransoms  & 3224    & 21100  & 15.28\cr
    D & Illegal BTC darknets  & 5838    & 109937 & 5.31\cr
    \bottomrule
    \bottomrule
    \end{tabular}
    \vspace{-0.3cm}
    \label{tab:addr_num}
    \end{center}
\end{table}

\subsection{Settings and Metrics}
\label{sec:metrics}
As our purpose is to detect malicious addresses as early as possible,
the model should detect them before the institution's daily settlement when the institutions may find the malice by themselves.
% Besides, after statistical analysis and the experimental result from the PU-Learning, $93.1\%$ addresses' behaviors become stable after the $24$-th hour (no further actions until one week later).
Therefore, our experiments focus on early illicit detection during the first day.
Although the experiments investigate the performance during the first day, our \emph{Evolve path Tracer} can work with an arbitrary timespan.
To evaluate the performance of our model, we get $24$ hours data with $1$ hour interval, and we average the evaluation metrics on all timesteps.
The property evolution and experiment environment can be found in Appendix~\ref{sec:appendix}.

The selected metrics are accuracy (Acc.), precision (Prec.), and recall (Rec.).
Besides, the model should predict correct labels fast to prevent economic loss earlier.
Also, due to data insufficiency, the model may predict conflict labels at different timesteps, thus confusing users. 
Therefore, we require the predictions to be consistent.
Similar to earliness and consistency requirements in \cite{25_}, we also design the early-weighted F1 score $F1^{E}$ and consistency-weighted score $F1^{C}$ as follows:
\begin{equation}
\begin{aligned}
\mathit{F1^{E}}& = \frac{\sum_{i=1}^{N}{F1_{i}/\sqrt{i}}}{\sum_{i=1}^{N}{1/\sqrt{i}}}, \\
\mathit{F1^{C}}& = \frac{\sum_{i=1}^{N-1}{{\sqrt{i}}\times{F1_{i}}\times{\mathbbm{1}_{y_c}(y_{i})}}}{\sum_{i=1}^{N-1}{{\sqrt{i}}}},
\end{aligned}
\end{equation}
where $i$ is the timestep, $y_c$ is the set of predictions where $sign((y_i-0.5)\times(y_{i+1}-0.5))>0$. The indicator function $\mathbbm{1}_{y_c}(y_{i})=1$ when $y_i \in y_c$. $F1_i$ is the $F1$ score of the prediction at timestep $i$.

\subsection{Effects of Uniform Path Length (RQ1)}
\label{sec:length_compare}
\begin{figure}
	\centering
	\vspace{-0ex}
	\includegraphics[width=1.\columnwidth, angle=0]{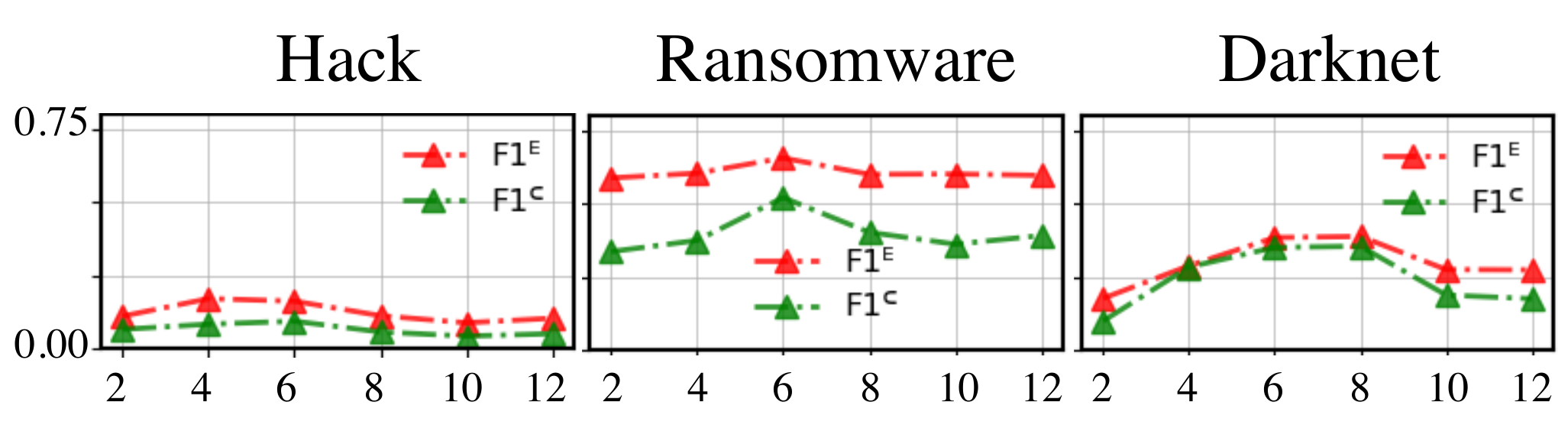}
	\vspace{-5ex}
	\caption{\emph{$F1^E$} and \emph{$F1^C$} of different uniform path lengths on three datasets.}
	\label{fig:lenghth_evolve}
	\vspace{-2ex}
\end{figure}

% As mentioned in Section \ref{sec:path_encoder}, we need to project asset transfer paths to the same length $L_u$ to encode them. 
We further analyze the effects of uniform path length with a simplified \textbf{AF/Path} model (using Address features and Asset Transfer Path features). We test 6 different path lengths on all three datasets (From $2$ to $12$ by the interval of $2$), as shown in Figure \ref{fig:lenghth_evolve}. 
% In this section, we only want to find the path length that maximizes the performance, we thus use simplified models for rapid testing which may not promise good performance.

A longer uniform path can preserve more information that contributes to better performance. So the model performs better as $L_u$ increases at the beginning. However, if the $L_u$ is longer than most asset transfer paths, the uniform path may introduce more redundant noise.
% Therefore, the model does not perform better when the path length is too large.

Since hack addresses get funds directly from the victim's account and need to transfer money as soon as possible, its asset transfer path is shorter. As shown in Fig~\ref{fig:lenghth_evolve}, the model achieves the best $F1^E$ and $F1^C$ scores when setting $L_u$ to 4 and 6, respectively. 
Considering both scores, the model performs best when $L_u$ equals $6$.
%
% Ransomware is malicious software that threatens the victims to pay a ransom fee. 
For Ransomware, the ransom demand comes with a deadline. Victims buy bitcoins from exchanges and transfer to criminals, thus slightly increasing the lengths of the asset transfer paths. As shown in Fig.~\ref{fig:lenghth_evolve}, the model performs best When the $L_u$ is $6$.
%
% A darknet is an overlay network within the Internet that can only be accessed with specific authorization. Thereby, 
For darknet, users could buy and sell illicit goods anonymously via them. Since platforms need to wait for the actions of buyers and sellers, there will be a longer asset transfer path. The model performs best When the $L_u$ is between $6$ to $8$. 
% However, setting $L_u$ to $6$, the model gets similar scores. 
Considering the model's scalability, in the actual experiment, we also set $L_u$ to $6$.
\begin{table*}
    \centering
    \fontsize{8}{11}\selectfont    %{字体尺寸}{行距}
    \caption{Scores of different prediction model. Evo-PT and Evo-PT (E) are our \emph{Evolve Path Tracer} with/wo ``Early Stop'' mechanism.
             Underline stands for best score in the group, Bold stands for best score in all groups.}
    \vspace{-2ex}
    \begin{tabular}{ccccccccccccccccc}
    \toprule
    \toprule
    \multirow{2}{*}{Type} & \multirow{2}{*}{\makecell{Model\\Name}} & \multicolumn{5}{c}{Hack} & \multicolumn{5}{c}{Ransomware} & \multicolumn{5}{c}{Darknet} \cr
    \cmidrule(lr){3-7} \cmidrule(lr){8-12} \cmidrule(lr){13-17} 
    &\quad  &$Acc.$ & $Prec.$ & $Rec.$ & $F1^E$ & $F1^C$ & $Acc.$ & $Prec.$ & $Rec.$ & $F1^E$ & $F1^C$ & $Acc.$ & $Prec.$ & $Rec.$ & $F1^E$ & $F1^C$  \cr
    \hline
    \multirow{3}{*}{\makecell{Machine \\ Learning}}
    &DT            &0.995	     &0.347	       &0.137	    &0.197	     &0.197        &0.955	    &0.736	      &0.432	    &0.545	     &0.545	       &0.982   	&0.448   	&0.152	    &0.227	       &0.227\cr
    &RF            &0.996	     &\ud{0.405}   &\ud{0.242}	&\ud{0.303}	 &\ud{0.303}   &0.955	    &0.735	      &\ud{0.436}	&0.547	     &0.547        &0.983   	&0.519   	&0.109   	&0.181	       &0.181\cr
    &XGB           &\udb{0.997}	 &0.347	       &0.137	    &0.197	     &0.197        &\udb{0.960}	&\udb{0.865}  &0.435	    &\ud{0.579}	 &\ud{0.579}   &\udb{0.985}	&\udb{0.790}	&\ud{0.191}	&\ud{0.308}	   &\ud{0.308}\cr
    \hline
    \multirow{5}{*}{\makecell{Sequen. \\ Deep \\ Learning}}
    &GRU           &0.928        &0.298        &0.438       &0.354 	     &0.354        &0.885	    &0.558	      &0.949	    &0.703	     &0.703	       &0.942	    &0.470	     &0.838   	  &0.603	   &0.603\cr
    &M-LSTM        &\ud{0.949}   &\ud{0.418}   &0.272	    &0.328       &0.333        &0.887	    &0.561	      &\udb{0.969}  &0.711	     &0.710        &\ud{0.951}	&\ud{0.520}	 &\udb{0.845} &\ud{0.642}  &\ud{0.645}\cr
    &CED	       &0.909        &0.265        &\ud{0.563}	&\ud{0.360}  &0.360        &\ud{0.909}  &0.617        &0.960	    &\ud{0.752}  &0.751	       &0.943	    &0.478	     &0.829	      &0.606	   &0.606\cr
    &SAFE          &0.918        &0.285        &0.438	    &0.271       &0.330        &\ud{0.909}  &0.616	      &0.963	    &\ud{0.752}  &0.752        &0.949	    &0.508	     &0.838	      &0.632	   &0.632\cr
    &TMIF          &0.913        &0.277        &0.560	    &0.356       &\ud{0.364}   &0.897       &\ud{0.623}	  &0.967	    &\ud{0.755}  &\ud{0.761}   &0.948	    &\ud{0.520}	 &0.840	      &0.638	   &0.639\cr
    \hline
    \multirow{3}{*}{\makecell{Addr. \\ Graph}}
    &GCN           &\ud{0.920}   &\ud{0.433}   &0.670	    &\ud{0.501}  &\ud{0.507}   &0.887       &0.564	      &0.936	    &0.700       &0.706        &\ud{0.942}  &\ud{0.459}  &0.613	      &0.525	   &0.524\cr
    &Skip-GCN      &0.917        &0.410        &0.690	    &0.443       &0.432        &0.903       &0.603        &0.935	    &0.729	     &0.735	       &0.941	    &\ud{0.459}  &0.629	      &\ud{0.531}  &0.530\cr
    &Evo-GCN       &0.893        &0.428        &\udb{0.749}	&0.427       &0.442        &\ud{0.906}  &\ud{0.613}   &\ud{0.944}   &\ud{0.736}  &\ud{0.746}   &0.941	    &\ud{0.459}  &\ud{0.633}  &0.530	   &\ud{0.533}\cr
    \hline
    \multirow{2}{*}{\makecell{TX. \\ Graph}}
    &Evo-PT    &0.963        &0.607        &\ud{0.739}  &0.664       &0.668        &0.938       &0.743        &\ud{0.869}   &0.799       &0.802        &0.963    	&0.624       &\ud{0.764}  &\udb{0.686} &0.686\cr
    &Evo-PT (E)  &\ud{0.969}   &\udb{0.650}  &0.731	    &\udb{0.689} &\udb{0.683}  &\ud{0.940}  &\ud{0.751}   &\ud{0.869}   &\udb{0.802} &\udb{0.807}  &\ud{0.964}  &\ud{0.625}  &0.754	      &\udb{0.686} &\udb{0.687}\cr
    \bottomrule
    \bottomrule
    \end{tabular}
    \vspace{-3ex}
    \label{tab:sota_compare}
\end{table*}

\subsection{Performance Comparison (RQ2)}
\label{sec:sota_compare}
\begin{table}
    \centering
    %\fontsize{10}{10}\selectfont  
    \caption{Scores of different ablation models on Hack (H), Ransomware (R), and Darknet (D). Ablation modules include Address Features (\textbf{AF}), Path features (\textbf{+Path}), Path Graph features (\textbf{+Graph}), and Evolve schemes (\textbf{+Evolve}).}
    \vspace{-1ex}
    \begin{tabular}{c|c|ccccc}
    \toprule
    \toprule
    &Model & $Acc.$ & $Prec.$ & $Rec.$ & $F1^E$ & $F1^C$ \cr
    \midrule %\rotatebox[origin=c]{90}
    \multirow{4}{*}{{H}}    &	AF	      &0.920	  &0.309	    &\bf{0.590}	&0.389	     &0.412	\cr
                            &	+Path	  &0.954      &0.537	    &0.546	    &0.509	     &0.538	\cr
                            &	+Graph	  &\bf{0.965} &\bf{0.686}   &0.476	    &0.545	     &0.559	\cr
                            &	+Evolve   &0.961      &0.606        &0.553	    &\bf{0.551}  &\bf{0.576}	\cr
    \midrule
    \multirow{4}{*}{{R}}    &	AF	      &0.911	  &0.710	    &0.632  	&0.626	     &0.628	\cr
                            &	+Path	  &0.929      &0.727	    &0.805	    &0.760	     &0.765	\cr
                            &	+Graph	  &0.927      &0.696	    &\bf{0.875}	&0.773	     &0.776	\cr
                            &	+Evolve   &\bf{0.937} &\bf{0.735}   &0.871	    &\bf{0.795} &\bf{0.798}	\cr
    \midrule
    \multirow{4}{*}{{D}}    &	AF	      &0.961	  &0.619	    &0.571	    &0.611	     &0.604	\cr
                            &	+Path	  &0.961      &0.611	    &0.693	    &0.649	     &0.650	\cr
                            &	+Graph	  &0.960      &0.586    	&\bf{0.804}	&0.678	     &0.678	\cr
                            &	+Evolve   &\bf{0.963} &\bf{0.626}   &0.758	    &\bf{0.685}  &\bf{0.685}	\cr
    \bottomrule
    \bottomrule
    \end{tabular}
    \vspace{-1ex}
    \label{tab:ablation}
\end{table}
\begin{figure}
	%\centering
	\vspace{-0ex}
	\includegraphics[width=1.\columnwidth, angle=0]{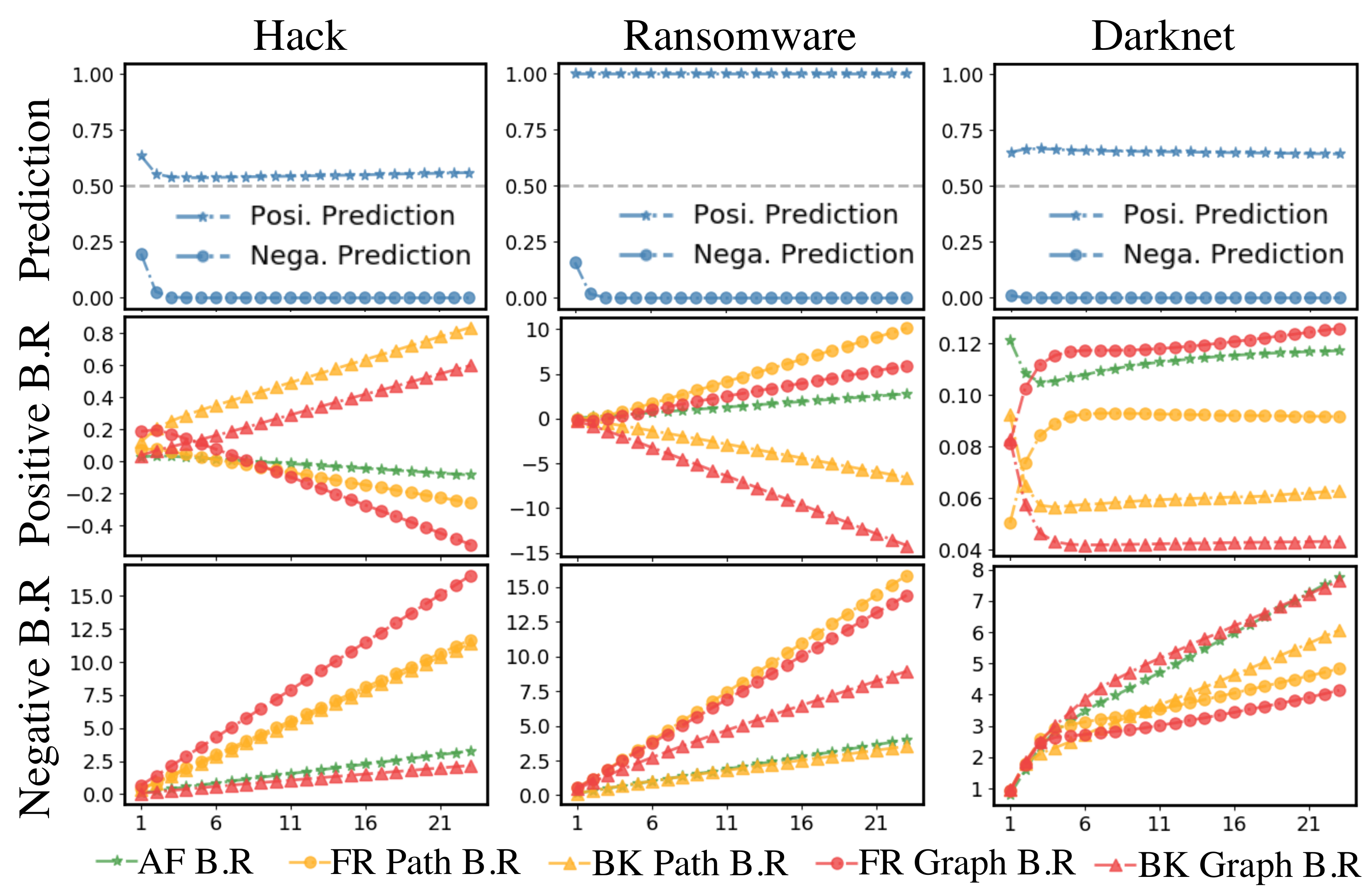}
	\vspace{-6ex}
	\caption{Prediction evolution of different address groups and corresponding average Benevolent Rates (B.R) of different features.}
	\label{fig:feature_contribution}
	\vspace{-4ex}
\end{figure}

To verify the effectiveness and versatility of our \emph{Evolve Path Tracer}, we first compare the most common machine learning models, then compare our encoder module with the encoder in other early detection models. At last, we also compare the address graph-based models.
The models are detailed in the appendix.
The main results for comparing all different methods are shown in Table \ref{tab:sota_compare}, and the major findings are summarized as follows:

(1) Our \emph{Evolve Path Tracer} outperforms most compared methods by a significant margin. Especially for early detection performance metrics F1-E and F1-C, \emph{Evolve Path Tracer} achieves the best performance under all three datasets. Compared to the second-best methods, \emph{Evolve Path Tracer} has an average increase of $14.54$\% on $F1^E$ and an average increase of $15.63$\% on $F1^C$. Besides, none of these methods can perform well on all three datasets, which justifies the effectiveness and versatility of our \emph{Evolve Path Tracer}.
Besides, the "Early Stop" mechanism accelerates the prediction speed and helps the model to discard subsequent noise.

(2) Traditional machine learning algorithms do not perform well on the three datasets.
It is difficult for decision-tree-based machine learning algorithms to encode the temporal shifts of the decision boundary for different  features~\cite{kdd_rv_1, kdd_rv_2}. 
Therefore, our model has an average improvement of $80.52$\% on $F1^E$ and $83.44$\% on $F1^C$ compared to the best decision-tree-based model.

(3) The sequential methods perform well on the three datasets. However, our \emph{Evolve Path Tracer} still has an average $21.82$\% improvement on $F1^E$ and $24.11$\% on $F1^C$ compared to the best model in this group. 
The first reason is that the inter-relationships among asset transfer paths can reveal specific transaction patterns (e.g., two addresses transfer money through multiple paths to avoid monitoring). 
In addition, since the transaction pattern evolves in the early stage, a static encoding module can hardly encode evolving information. 

(4) Compared with Address Graph methods, \emph{Evolve Path Tracer} has average increases of $22.74$\% and $23.36$\% on $F1^E$ and $F1^C$ respectively. 
Among them, Evolve-GCN performs the best in most datasets, which verifies the fast-evolving of early transaction networks. 
However, as \cite{34_} implies, the address GCN may lead to Over-Smoothing issues and the dilution of the minority class. 
Moreover, as most neighbors of malicious nodes are victims or shadow addresses, the Address Graph models do not perform well.
To avoid these problems, in \emph{Evolve Path Tracer}, we set vertices as transactions to utilize the relevant address information in a "safer" way.

\subsection{Ablation Study (RQ3)}
\label{sec:ablation_compare}
As shown in Table \ref{tab:ablation}, \textbf{AF} performed poorly on the three datasets. Because many benevolent addresses (change address, ICO, and legal market addresses) behave similarly to these malicious addresses. As shown in Fig.~\ref{fig:feature_contribution}, the AF benevolent rates for most negative samples do not exceed 2.5.
The introduction of asset transfer path features significantly improves the performance. 
As shown in Fig.~\ref{fig:feature_contribution}, for the Hack addresses, the forward transaction signal is more important than the backward one because the Hack address will transfer the funds faster and more centralized.
Ransomware and Darknet addresses usually require victims or buyers to transfer funds according to certain conditions, the backward information is thus more critical.

Comparing \textbf{+Path} and \textbf{+Graph}, 
by encoding the paths' interrelationships, the model gives predictions based on transaction patterns rather than the fluctuation of a single path.
As shown in Fig.~\ref{fig:feature_contribution}, the prominent signals of malicious nodes are enhanced by introducing path graphs. 
In the cryptocurrency transaction network, the model should be able to handle the differences between various types of addresses dynamically.
By comparing the performance differences between \textbf{+Graph} and \textbf{+Evolve}, we found that this Evolve mechanism is necessary. 
\textbf{+Graph} only performs well if the address has a shorter life span, and these addresses will be discarded after the first few transactions. 
However, for other addresses with longer lifetimes, \textbf{+Evolve} can better reflect changes in the transaction patterns of these addresses, resulting in better performance.

\subsection{Scalability and Dynamical Prediction (RQ4)}
\label{sec:scalability}
\noindent\textbf{Feature Preparation Time Cost}.
When a new block appears, users will usually monitor the addresses that have transactions with them. 
Those new and large-volume addresses are likely to participate in dangerous activities. 
Therefore, we randomly selected $1,000$ blocks (from the first block of $2018$ to the first block of $2022$) and collected the daily BTC price during this period. 
We filter out transactions lower than \$$10,000$ and retrieve addresses with a lifespan of less than one week.
We prepare every address's data for the first $24$ hours, the time cost is illustrated in Table~\ref{tab:time_cost}.
\begin{table}
\centering
\fontsize{10}{12}\selectfont 
\caption{Time cost of different input data, including Block Number, Transaction Number, Address Number), Address Feature, Asset Transfer Path, Path Graph, and Address Graph.}
\vspace{-1ex}
\begin{tabular}{cccccccccc}
\toprule
\toprule
\#Blk &\#TX &\#Addr &A-Feat &Path &P-G &A-G\cr
\midrule 
1,000 & 306,258  & 194,310    & 175s  & 7.7h     &6.9h   &14.3h\cr
Avg.     & 306      & 194        & 0.2s & 27.6s     &24.8s   &51.4s\cr
\bottomrule
\bottomrule
\end{tabular}
\vspace{-4ex}
\label{tab:time_cost}
\end{table}
During each interval (\textbf{1} hour is about $6$ blocks), we need to monitor about $1,166$ new addresses, which will only cost \textbf{5} minutes. 
Moreover, as shown in Table \ref{tab:time_cost}, our time cost resembles address graph preparation, but we can collect information much further than $2$ hops.

\begin{figure}
	%\centering
	\vspace{-0ex}
	\includegraphics[width=1.\columnwidth, angle=0]{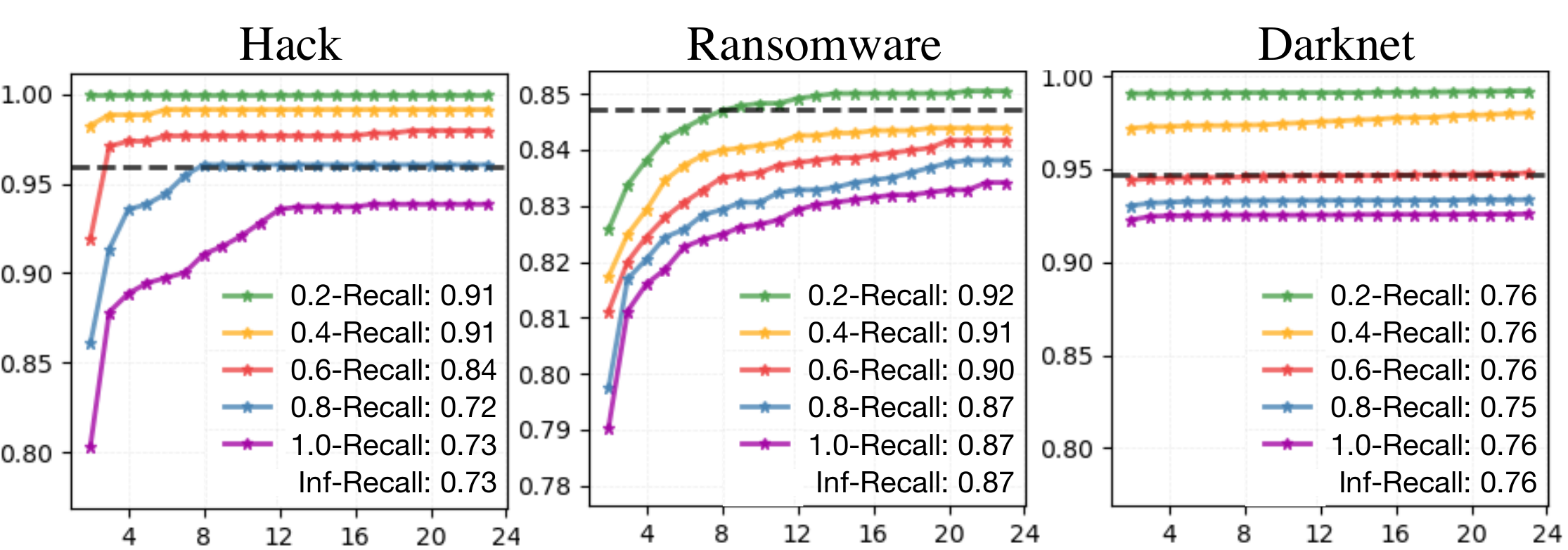}
	\vspace{-4ex}
	\caption{Skip Ratio evolution and $Recall$ scores with different thresholds. The gray line is the benevolent ratio of each dataset.}
	\label{fig:early_stop}
	\vspace{-5ex}
\end{figure}

\noindent\textbf{Scalability of Early Stop}.
We plot the skip ratios with different thresholds to justify the scalability.
As shown in Fig.~\ref{fig:early_stop}, all models can filter out most ($80$\%) addresses by the fourth hour.
The mechanism improves the model's Scalability significantly.
Moreover, a reasonable threshold helps the model to discard subsequent noise and improve the performance, as mentioned in Section~\ref{sec:sota_compare}.
A lower threshold means a faster prediction speed.
\emph{Inf} means no ``Early Stop'' mechanism, the skip ratio is thus always $0$.

There is a concern about missing malicious addresses as we decrease the threshold, 
which then decreases the model's \emph{Recall} scores.
However, as shown in Fig.~\ref{fig:early_stop}, compared to the model without ``Early Stop'' (labeled as ``Inf''), the model has better \emph{Recall} scores as we decrease the threshold.
This is because our model can predict the most benevolent addresses in the early hours.
Removing them from the monitoring list can avoid subsequent noise, which improves the model’s \emph{Recall} scores. 
Therefore, our \emph{Evolve Path Tracer} has a faster prediction speed without missing malicious addresses.

%------------------------------------------------------------------------
\section{Conclusion and Future Work}
\label{sec:conclusion}
In this paper, we present \emph{Evolve Path Tracer}, a novel framework for early malicious address detection. 
We first propose Asset Transfer Paths and encode them with \emph{Evolve Path Encoder LSTM}. 
The asset transfer paths exhibit high versatility in monitoring various transaction patterns in the early stage. 
Then the \emph{Evolve Path Graph GCN} is built to encode corresponding path graphs. 
The graphs capture the interrelation among the paths. 
In particular, all modules are evolving dynamically to encode the dynamics of paths' structure and inter-relation. 
Finally, we implement \emph{Hierarchical Survival Predictor} with \emph{Consistency Loss Function} to achieve better prediction performance, higher consistency, and excellent scalability.
The model is comprehensively evaluated on three datasets. 
Extensive ablation studies explain the mechanisms behind the effectiveness and excellent scalability. 
%In future work, we would like to extend \emph{Evolve Path Tracer} to malicious address detection in other crypto-currency platforms and traditional financial domains.

%------------------------------------------------------------------------
\section{Acknowledgments}
This research is supported by the National Research Foundation, Singapore under its Emerging Areas Research Projects (EARP) Funding Initiative, the NSFC (72171071, 72271084, 72101079) and the Excellent Fund of HFUT (JZ2021HGPA0060). Any opinions, findings and conclusions or recommendations expressed in this material are those of the author(s) and do not reflect the views of National Research Foundation, Singapore.

%------------------------------------------------------------------------
\bibliographystyle{ACM-Reference-Format}
\bibliography{sample-base}

\clearpage
\appendix
\section{Supplementary Material}
\label{sec:appendix}
\subsection{Reproducibility}
% We release Evolve Path Tracer on \textcolor{blue}{GitHub\footnote{https://github.com/Cranooooooo/ADS-Demo}}. 
We first download full-node BTC raw data with Bitcoin Core. The whole data size is about 500GB. After downloading all blocks before the 700,000th block, we parse all data by Blocksci for querying block, transaction, and address index. The parsed data size is about 400GB. For each address in our dataset, we prepare its asset transfer paths for the transactions during the first 24 hours. The process was executed on AMD Ryzen 9 3900X Processor with 64GB of memory. We implement Evolve Path Tracer in Pytorch and Geometric. All experiments are conducted on a single NVIDIA RTX 2080TI with 11GB memory.
The dimension of hidden layers in our model is 32, with a total of 919203 (~0.9M) parameters. The size of the model's checkpoint is only 3.57 MB. The training was performed for a maximum of 40 epochs. Early stopping is applied if the best performance didn't update in the latest five epochs. The training and testing time cost for each dataset is shown in Table.~\ref{tab:time_cost}:

\begin{table*}[]
\centering
\fontsize{8}{9}\selectfont 
\caption{Key components and module description. TX stands for the transaction. BK and FR stand for Backward and Forward.}
\begin{tabular}{cc}
\toprule
Name(Notation) &Description  \cr
\midrule 
Influence TX pair ($j \rightarrow i$) & A certain portions of TX i's BTCs come from TX j \cr
            \cmidrule(l{0em}r{0em}){2-2} 

Trust TX pair ($i \rightarrow j$) & A certain portions of TX i's BTCs flow to TX j \cr
            \cmidrule(l{0em}r{0em}){2-2} 

BK Asset Transfer Path ($j_n \rightarrow \cdots \rightarrow i$) & Build the Influence TX pairs iteratively and link them to form a path  \cr
            \cmidrule(l{0em}r{0em}){2-2} 

FR Asset Transfer Path ($i \rightarrow \cdots \rightarrow j_n$) & Build the Trust TX pairs iteratively and link them to form a path \cr
            \cmidrule(l{0em}r{0em}){2-2} 

BK/FR Path Graph   & BK/FR Asset Transfer paths share the same source/destination TX are grouped to form a graph \cr
            \cmidrule(l{0em}r{0em}){2-2} 

\midrule 
 T-1 LSTM   & LSTM to encode temporal information of address features \cr
            \cmidrule(l{0em}r{0em}){2-2} 

E-1/2 LSTM   & An Evolve Path Encoder LSTM for encoding BK/FR Asset Transfer Path to Path feature \cr
            \cmidrule(l{0em}r{0em}){2-2} 

T-2/3 LSTM   & LSTM to encode temporal information of BK/FR Path feature \cr
            \cmidrule(l{0em}r{0em}){2-2} 

BK/FR-Graph Encoder  & An Evolve Path Graph GCN for encoding BK/FR Path Graph to Graph feature \cr
            \cmidrule(l{0em}r{0em}){2-2} 

T-4/5 LSTM   & LSTM to encode temporal information of BK/FR Graph feature \cr

\bottomrule
\label{tab:module_explain}
\vspace{-2ex}
\end{tabular}
\end{table*}

\begin{table}
    \caption{Time Cost}
    \vspace{-0.3cm}
    \begin{center}
    \fontsize{10}{12}\selectfont
    \renewcommand{\arraystretch}{1.2}
    \begin{tabular}{ccccc}
    \toprule
    \toprule
    Type  & Train & Test & Time-Span & Sample Num. \cr
    \midrule
    H & 0.6 hours & 0.6 mins    & 24 hours & 6,884\cr
    R & 1.1 hours & 2.3 mins    & 24 hours & 24,324\cr
    D & 7.6 hours & 16.3 mins   & 24 hours & 115,775\cr
    \bottomrule
    \bottomrule
    \end{tabular}
    \vspace{-0.3cm}
    \label{tab:time_cost}
    \end{center}
\end{table}

\begin{figure*}
	%\centering
	\vspace{-0ex}
	\includegraphics[width=2.\columnwidth, angle=0]{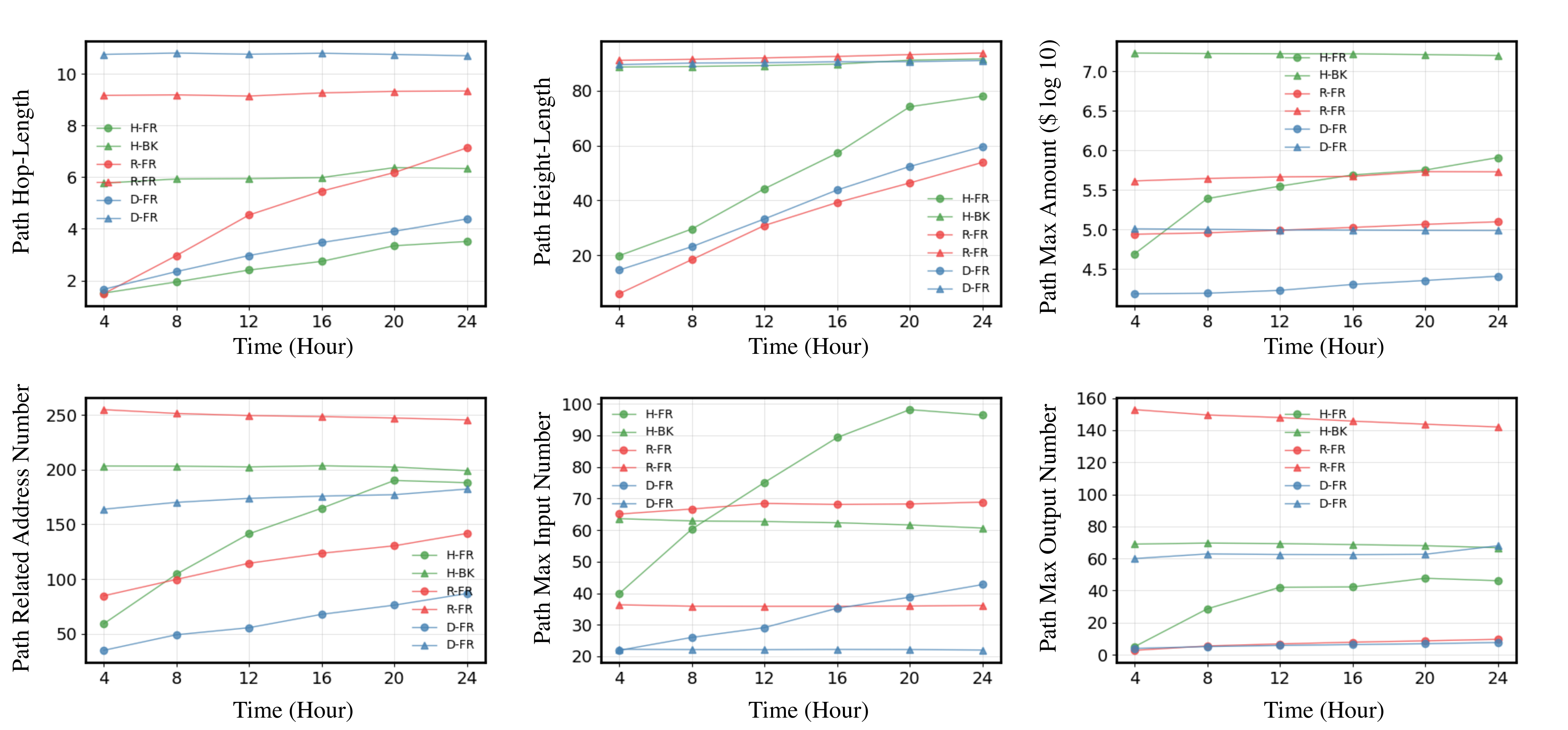}
	\vspace{-2ex}
	\caption{Asset transfer path's statistical properties of different malicious addresses under the backward and forward direction.}
	\label{fig:path_prop}
	\vspace{-0ex}
\end{figure*}

\subsection{Label Acquisition}
To get the labels for three different illicit activities (Hack, Ransomware, and Darknet), 
we performed a manual search on public forums and datasets, such as Bitcointalk forum\footnote{https://bitcointalk.org/}, Reddit, WalletExplorer\footnote{https://www.walletexplorer.com} and several prior studies \cite{19_, 20_, 14_}. 
Negative (Regular) addresses are collected in the same method as ~\cite{14_, 21_}.
We set the activation threshold as $0.01$ to prepare the asset transfer path. One can set a smaller threshold depending on the operating environment. 

\subsection{Address Features}
We use the following features to characterize an address at a specific timestamp.
\begin{itemize}
    \item the current balance of the address
    \item the number of receive (spend) transactions, 
    \item the ratio of receiving (spend) transactions number, 
    \item the maximum receive (spend) transactions number, 
    \item the life span of the address, 
    \item address active rate.
\end{itemize}

\subsection{Transaction Features}
We use the following features to characterize a transaction, which is the component of every asset transfer path.
\begin{itemize}
    \item the height interval to the path source,
    \item the influence (trust) score with the previous transaction,
    \item the input amount of the previous transaction, 
    \item the transaction fee,
    \item the total amount (resp. max, min, avg, and var) of all receive (spend) transactions, 
    \item the number of receive (spend) transactions. 
\end{itemize}

\subsection{Preparation of Asset Transfer Path}
Algo. \ref{alg:bk_path} gives the detail to prepare \emph{Backward Asset Transfer Paths} that reveal where $j$ obtains the asset.
The pipeline to construct \emph{Forward Asset Transfer Path} is similar to \emph{Backward Asset Transfer Path}. The only difference is the tracing direction.
The essence of each node is a transaction, so we use a sequence of transaction features to represent an asset transfer path.
 \begin{algorithm}
 \caption{Backward Path Preparation}
 \label{alg:bk_path}
  \DontPrintSemicolon
  \SetKwInOut{Input}{input}
  \SetKwInOut{Output}{output}
  \Input{Initial Output Tx $j^o$, Threshold $\theta$, Time Span $T_{Span}$.}
  \Output{Backward Path Set $P$.}
 
 \nl Initialize Backward Path Set: $P \gets \{[-,1,j^o]\}$;\;
 \nl Initialize Previous hop recorder: $P_{pre} \gets \{[-,1,j^o]\}$;\;
 \nl Initialize Ending Flag: $F_{end} \gets False$;\;
 \nl $j^o$'s Time: $T_{j^o} \gets $ Time of $j^o$;\;

  \nl \While{$F_{end} \ne True$}{
  \nl Current hop recorder $P_{now} \gets \{\}$;\;
  \nl $F_{end} \gets True$;\;
  \nl  \For {$p$ in $P_{pre}$}{
  \nl        $j \gets $ Output Tx $p[2]$;\;
  \nl        $I \gets $ Input Tx Set of $j$;\;
  \nl        \For {$i$ in $I$}{
  \nl                  $Prop_{i} \gets Amt_{i}/Amt_{I}$;\;
  \nl              $Score_{i} \gets Prop_{i}*p[1]$;\;
  \nl              $T_{i} \gets $ time of $i$;\;
  \nl              \If {($Score_{i} \ge \theta$ and $T_{j^o} - T_{i}\le T_{Span}$)}{
  \nl                     Append $[j, Score_{i}, i]$ to $P_{now}$;\;
  \nl                     $F_{end} \gets F_{end}$ $\&\&$ $False$;\;}}}
\nl $P_{pre} \gets P_{now}$;\;
\nl $P \gets P \cup P_{pre}$;\;
}
\nl \Return{$P$} 
\end{algorithm}
\vspace{-0ex}

\subsection{Baseline Models}
We give details of our baseline methods from two related tasks:

\noindent\textbf{Malicious Detection in Cryptocurrency}.
We compare Evolve Path-Tracer with several models for malicious address detection in cryptocurrencies. 
For decision tree models, we use address features and path statistic features as the feature set.
For GCN models, at each time step, we get the addresses' embedding after two graph convolutional layers as implemented in ~\cite{7_}.
Then, we feed the embeddings into a sequential model for prediction.
\begin{itemize}
\item
\textbf{Decision Tree}~\cite{32_, 31_} utilize Decision Tree for identifying these malicious addresses. 
\item
\textbf{Random Forest}~\cite{31_} utilize Decision Tree for identifying these malicious addresses. 
\item
\textbf{XGB}~\cite{33_, 31_} predict the type of a yet-unidentified entity with Gradient Boosting based algorithms. 
\item
\textbf{GCN}~\cite{7_} encodes the objective address based on its transaction address graph.
\item
\textbf{Skip-GCN}~\cite{7_} inserts a skip connection between the intermediate embedding and the input node features.
\item
\textbf{Evolve-GCN}~\cite{7_} updates GCN weights with an RNN module.
\end{itemize}

\noindent\textbf{Early Rumor Detection on Social Media}. For these sequential models, we build an extra path LSTM encoder for a fair comparison. We concatenate address features with the path-encoder output and feed them into the sequential prediction model.
\begin{itemize}
\item
\textbf{GRU}~\cite{27_} is a typical neural network for sequence modeling. At each time split, previous hidden state and current summation features are fed into the GRU unit to predict the labels for the given addresses.
\item
\textbf{M-LSTM}~\cite{23_} adopts LSTM for every kind of feature to generate its own temporal features at each timestamp. Here we build three LSTM models for Address Features, Forward Paths, and Backward Paths.
\item
\textbf{CED}~\cite{25_} also uses GRU for sequence modeling, it proposes the concept of “Credible Detection Point,” making it possible to make predictions as early as possible dynamically.
\item
\textbf{SAFE}~\cite{22_} adopts survival probability as the prediction. Instead of predicting the labels directly, it generates hazard rates for the survival models. The positive samples (Malicious Addresses) should die out fast, while the negative samples (Regular Addresses) should stay alive.
\item
\textbf{TMIF}~~\cite{35_} is a transformer-based multi-modal mode to capture the dependency relationship between multi-modal content. Here we set the modal to be Address-Modal and Path-Modal.
\end{itemize}

\end{document}